\documentclass{SCIS2025}

\usepackage{graphicx}
\usepackage{amstext}
\usepackage{amsmath}
\usepackage{hyperref}
\usepackage{cleveref}
\Crefname{section}{§}{§§}
\Crefname{section}{§}{§§}
\usepackage{threeparttable}
\usepackage{makecell}
\usepackage{booktabs}
\usepackage{stfloats}
\usepackage{multirow}
\usepackage{bbding}
\usepackage{rotating}
\usepackage{tabularx}
\usepackage{colortbl}
\usepackage{wrapfig}
\usepackage{arydshln}
\usepackage{xcolor}
\newfloat{figtab}{htb}{fgtb}
\makeatletter
  \newcommand\figcaption{\def\@captype{figure}\caption}
  \newcommand\tabcaption{\def\@captype{table}\caption}
\makeatother

\def \multi {\textsc{Multi}}
\def \multihard {\textsc{Multi-Elite}}
\def \multikn {\textsc{Multi-Extend}}

\def \Multi {M\scalebox{0.8}[0.8]{\MakeUppercase{ULTI}}}
\def \Multihard {M\scalebox{0.8}[0.8]{\MakeUppercase{ULTI}}-E\scalebox{0.8}[0.8]{\MakeUppercase{lite}}}
\def \Multikn {M\scalebox{0.8}[0.8]{\MakeUppercase{ULTI}}-E\scalebox{0.8}[0.8]{\MakeUppercase{xtend}}}

\def \V {\scriptsize\Checkmark}

\newcommand{\U}[1]{\underline{#1}}
\newcommand{\B}[1]{\textbf{#1}}

\newcommand{\R}[1]{\textcolor[rgb]{0.75, 0.25, 0.25}{\B{#1}}}
\newcommand{\G}[1]{\textcolor[HTML]{578c2e}{\B{#1}}}
\definecolor{rmbblue}{HTML}{b5aabd}

\begin{document}
\ArticleType{RESEARCH PAPER}
\SpecialTopic{Special Topic: Large Multimodal Models}
\Year{2025}
\Month{October}
\Vol{68}
\No{10}
\DOI{10.1007/s11432-024-4602-x}
\ArtNo{200107}
\ReceiveDate{15 May 2024}
\ReviseDate{19 December 2024}
\AcceptDate{22 July 2025}
\OnlineDate{28 September 2025}
\AuthorMark{Zhu Z C}
\AuthorCitation{Zhu Z C, Xu Y, Chen L, et al}

\title{MULTI: Multimodal Understanding Leaderboard with Text and Images}{MULTI: Multimodal Understanding Leaderboard with Text and Images}


\author[1,3,4]{Zichen ZHU}{}
\author[1,3]{Yang XU}{}
\author[1,2,3,5]{Lu CHEN}{{chenlusz@sjtu.edu.cn}}
\author[1,3]{Jingkai YANG}{}
\author[1,3]{Yichuan MA}{}
\author[1,3]{\\Yiming SUN}{}
\author[1,3]{Hailin WEN}{}
\author[1,3]{Jiaqi LIU}{}
\author[1,3]{Jinyu CAI}{}
\author[1,3]{\\Yingzi MA}{}
\author[1,3]{Situo ZHANG}{}
\author[1,3]{Zihan ZHAO}{}
\author[1,3]{Liangtai SUN}{}
\author[1,3,5]{Kai YU}{{kai.yu@sjtu.edu.cn}}

\AuthorMark{Zhu Z C}

\AuthorCitation{Zhu Z C, Xu Y, Chen L, et al}


\address[1]{X-LANCE Lab, School of Computer Science, Key Laboratory of Artificial Intelligence\\Ministry of Education, Shanghai Jiao Tong University, Shanghai {\rm 200240}, China}
\address[2]{Shanghai Innovation Institude, Shanghai {\rm 200030}, China}
\address[3]{Jiangsu Key Lab of Language Computing, Suzhou {\rm 215123}, China}
\address[4]{College of Computing and Data Science, Nanyang Technological University, Singapore {\rm 639798}, Singapore}
\address[5]{Suzhou Laboratory, Suzhou {\rm 215123}, China}

\abstract{
The rapid development of multimodal large language models (MLLMs) raises the question of how they compare to human performance. While existing datasets often feature synthetic or overly simplistic tasks, some models have already surpassed human expert baselines. In this paper, we present \multi, a Chinese multimodal dataset derived from authentic examination questions. Comprising over 18,000 carefully selected and refined questions, \multi~evaluates models using real-world examination standards, encompassing image-text comprehension, complex reasoning, and knowledge recall. Additionally, we also introduce \multihard, a 500-question selected hard subset, and \multikn~with more than 4,500 external knowledge context pieces for testing in-context learning capabilities. Our evaluation highlights substantial room for MLLM advancement, with Qwen2-VL-72B achieving a 76.9\% accuracy on \multi~and 53.1\% on \multihard~leading 25 evaluated models, compared to human expert baselines of 86.1\% and 73.1\%. \multi~serves not only as a robust evaluation platform but also paves the way for the development of expert-level AI. Details and access are available at \url{https://OpenDFM.github.io/MULTI-Benchmark}.

}


\maketitle


    \begin{figure}[!h]
        \centering
        \includegraphics[width=0.7\textwidth,trim=34 215 35 215,clip]{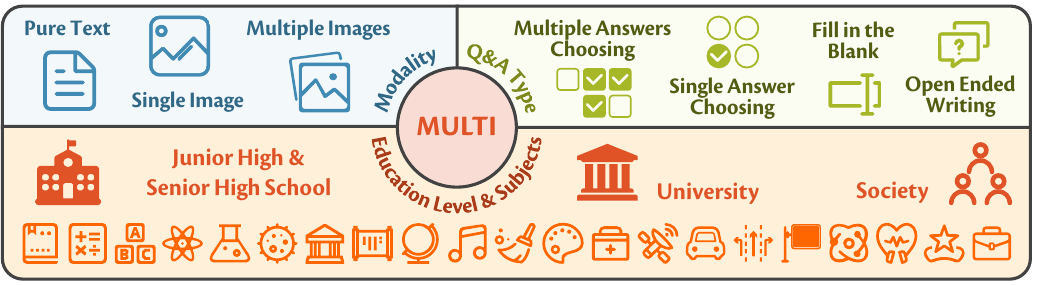}
        \caption{\textbf{\Multi~is a large-scale Chinese multimodal benchmark across multiple domains and aspects.}}
        \label{figure:overview}
        \vspace{-0.5cm}
    \end{figure}

\section{Introduction}

The rapid development of large-scale language models (LLMs)~\cite{sun2023moss,chen2022dialogzoo,qwen2,cai2024internlm2,chatgpt,gpt4,geminiteam2023gemini,touvron2023llama} has brought about remarkable progress in natural language processing (NLP) and related fields~\cite{lan2024depression,han2024ibsen,zhao2025developing,xurejection,xureducing,zhang2025reasoning}. However, human communication and understanding extend far beyond just language, encompassing various other forms of information such as images, mathematical formulas, graphs, tables, and diagrams. These visual modalities play an essential role in fields like science and engineering, where they are crucial for conveying complex ideas. Consequently, there has been increasing interest in developing Multimodal Large Language Models (MLLMs)~\cite{du2022glm,viscpm,cllava,bai2023qwen,yi,chen2023internvl,chen2024far,yao2024minicpm,wang2024qwen2,gpt4v,geminiteam2023gemini,gpt4omini,gpt4o,claude3}, that can process and generate across different modalities, including visual ones, and perform tasks that require cross-modal reasoning~\cite{ma2024dolphins,zhu2025moba,zhao2024chemdfm}. All these rapid advancements raise the question of how a model compares to a real student.

Evaluating MLLMs presents unique challenges, as current benchmarks~\cite{lu2022learn, li2023seed} are either too narrow in scope, focusing on natural scene images, or too simplistic to effectively assess the capabilities of modern models. Many existing scientific benchmarks~\cite{sun2023scieval, huang2023ceval} rely on multiple-choice questions with a single correct answer, which may not adequately test a model's comprehension or reasoning skills. As a result, models can sometimes arrive at superficial solutions without fully understanding the problem. A more robust, detailed, and multi-scale dataset is necessary to evaluate MLLMs in a more comprehensive manner under diverse real-world conditions. Additionally, while most existing benchmarks are in English, the rapid progress of Chinese multimodal models has highlighted the need for a Chinese multimodal benchmark that includes both Chinese text and visual content, offering unique challenges to the field.

In this paper, we introduce \multi, a new multimodal benchmark for evaluating LLMs on cross-modal understanding tasks. Our primary goal is to evaluate MLLMs across a broad range of tasks that a typical Chinese student would encounter throughout their academic progression, as shown in \Cref{figure:overview}. The \multi~benchmark, which stands for \textbf{M}ultimodal \textbf{U}nderstanding \textbf{L}eaderboard with \textbf{T}ext and \textbf{I}mages, consists of 18,430 carefully curated questions derived from real-world educational and online sources. These questions span multiple scientific domains, including mathematics, physics, and computer science, and feature a variety of question formats, such as multiple-choice, fill-in-the-blank, and open-ended questions. Importantly, \multi~also includes driving and administrative aptitude tests, which are significant in the Chinese context and offer further challenges in cross-modal reasoning and generation.

\begin{wrapfigure}[30]{r}{0.45\textwidth}
        \centering
        \vspace{-0.5cm}
        \includegraphics[width=0.44\textwidth,trim=10 130 10 10,clip]{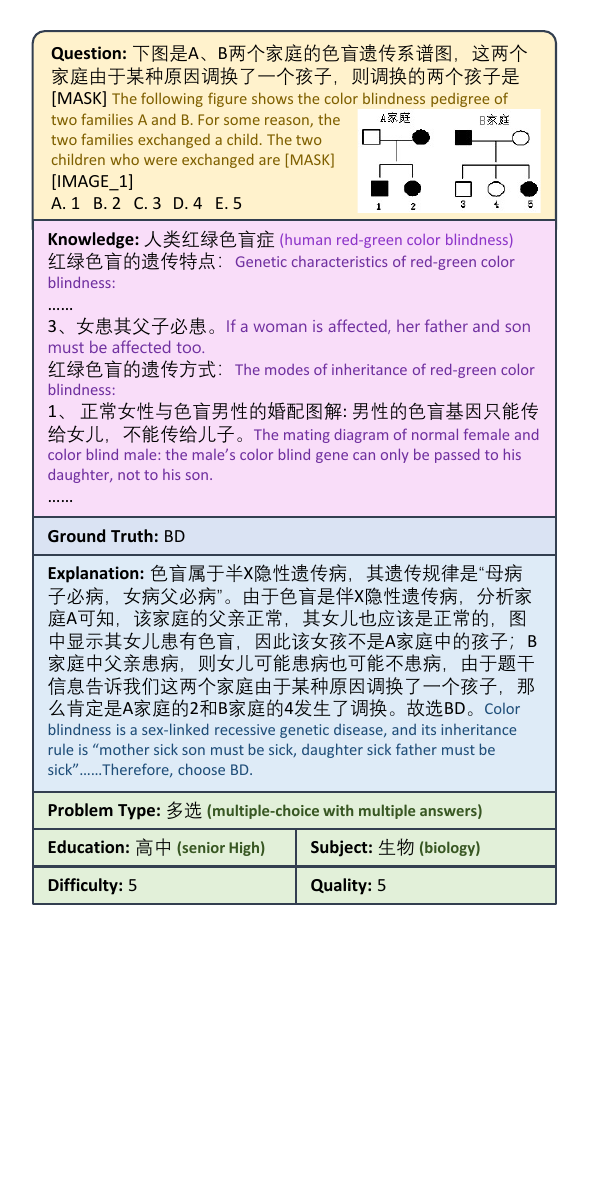}
        \caption{\textbf{An example of \Multi.} English translations of Chinese text are shown for better readability. The markdown format remains as it is.}
        \label{figure:example}
    \end{wrapfigure}

To push the boundaries of MLLM evaluation, we also introduce two specialized subsets within \multi: \multihard, which includes 500 tough questions designed to test the limits of current MLLMs, and \multikn, a set of 4,596 external knowledge pieces that assess models' abilities in in-context learning and knowledge transfer. These subsets provide deeper insights into the strengths and weaknesses of MLLMs, driving further research and model improvement. Sample questions from \multi~are presented in \Cref{figure:example}, and additional details are available in~\ref{sec:app-data} and~\ref{sec:app-case}.

We conduct extensive experiments on \multi, including both multimodal and single-modality LLMs, including 14 open-sourced and 11 close-source models. Our results show that while multimodal LLMs have made significant progress, they still lag behind human expert performance on many aspects of the \multi~benchmark. Key challenges remain in cross-modal alignment, logical reasoning, mathematical computation, and image comprehension. In particular, the \multihard~subset proves especially challenging, with the best-performing model, Qwen2-VL-72B, achieving only a 50.0\% accuracy, and many models performing near random levels. This highlights a substantial space for improvement in current MLLMs.

In summary, the contributions of this paper are as follows:
\begin{itemize}
    \item We introduce \multi, a comprehensive and challenging multimodal benchmark designed to evaluate Chinese scientific knowledge tasks for MLLMs.
    \item We present the \multihard~and \multikn~subsets to evaluate models' reasoning capabilities and in-context learning abilities, providing a more nuanced assessment of MLLMs.
    \item We conduct detailed experiments with 25 state-of-the-art multimodal and single-modality models, offering both qualitative and quantitative insights into their performance on \multi. We also conducted a bunch of ablation studies to draw further conclusions.
    \item We conduct human evaluations on 1k questions and give error analysis on 1.3k responses from models.
\end{itemize}

    \section{Related Work}

    \paragraph{Multimodal Large Language Models~(MLLMs).}

    With advancements in aligning features across multiple modalities, like CLIP~\cite{Radford2021LearningTV} and ALBEF~\cite{Li2021AlignBF}, recent studies have explored projecting vision features into the latent space of LLMs, aiming to enhance their capabilities of comprehending visual information. For example, BLIP-2~\cite{li2023blip} pioneers this approach by employing Q-Former to translate image features into text representations. Following this, LLaVA~\cite{liu2023visual}, MiniGPT-4~\cite{zhu2023minigpt}, and InstructBLIP~\cite{Dai2023InstructBLIPTG} have introduced visual instruction tuning to bolster the capability of MLLMs of following instructions. Our primary focus is on the proficiency of MLLMs in comprehending instructions in Chinese, which are divided into two main branches: open-source models, which typically build upon existing Chinese LLMs or are fine-tuned on Chinese instruction datasets, examples of which include Chinese-LLaVA~\cite{cllava}, VisualGLM~\cite{du2022glm}, VisCPM~\cite{viscpm}, Qwen-VL~\cite{bai2023qwen}, InternVL~\cite{zhang2023internlm}, Yi-VL~\cite{yi}; and closed-source models, which are often highly powerful, multi-lingual systems such as GPT-4V(ision)~\cite{gpt4v} and Gemini~\cite{geminiteam2023gemini}. In this paper, we intend to evaluate these models across a range of scientific fields on the \multi~benchmark, offering an extensive assessment and guidance for the onward trajectory of Chinese MLLMs.


    \paragraph{Benchmarks for MLLMs.} In assessing MLLMs, traditional methods primarily rely on established vision-language~(VL) benchmark datasets. Renowned benchmarks such as VQA~\cite{antol2015vqa,goyal2017making}, OK-VQA~\cite{marino2019ok}, GQA~\cite{hudson2019gqa}, and MSCOCO~\cite{lin2014microsoft} are tailored to specific VL tasks like image captioning, open-domain visual question answering, and visual reasoning. While the evaluation based on standard benchmark datasets yields significant insights into MLLMs' capabilities, these approaches may not entirely capture their comprehensive intelligence in real-world scenarios. Therefore, a diverse array of benchmarks has been developed to examine MLLMs in dealing with various tasks in the real world. Benchmarks like LLaVA-Bench~\cite{liu2023visual}, MMBench~\cite{liu2023mmbench}, MM-VET~\cite{yu2023mm}, TouchStone~\cite{bai2023touchstone}, MLLM-bench~\cite{ge2023mllm}, and SEED-Bench\cite{li2023seed,li2023seed2}, for instance, leverage GPT to evaluate the relevance and helpfulness of human-like long responses in the reality. POPE~\cite{li2023evaluating} and HallusionBench~\cite{liu2023hallusionbench} introduce various analytical criteria for the holistic evaluation of MLLMs' hallucinations. Furthermore, M3Exam~\cite{zhang2023m3exam}, SciGraphQA~\cite{li2023scigraphqa}, MathVista~\cite{lu2023mathvista}, AGIEval~\cite{zhong2023agieval}, and MMMU~\cite{yue2023mmmu} consider MLLMs as experts to extend the evaluation scope by incorporating advanced perception and reasoning within domain-specific knowledge, for example, scientific questions and driving tests. The works most related to us are M3Exam, ScienceQA, SciEval~\cite{sun2023scieval}, and C-Eval~\cite{huang2023ceval}. Our approach distinguishes itself by offering a broader spectrum of question types compared to the first two and supports a multimodal evaluation in contrast to the last two.

    \section{The \Multi~Benchmark}

    We propose \multi, a \texttt{M}ultimodal \texttt{U}nderstanding \texttt{L}eaderboard with \texttt{T}ext and \texttt{I}mages, which can serve as a challenging and diverse benchmark for the MLLM community. In \Cref{sec:dcp}, we provide an overall introduction to the data construction process, with more details provided in \ref{sec:app-data}. The detailed statistics are provided in \Cref{sec:stat}. In \Cref{sec:3.3} and \Cref{sec:3.4} we introduce \multihard~and \multikn. We also give a comparison between \multi~and other knowledge-based benchmarks in \Cref{sec:3.5}.

    \subsection{Data Construction Process}
    \label{sec:dcp}

\begin{figure}[!th]
        \centering
        \includegraphics[width=1.0\textwidth,trim=0 195 0 180,clip]{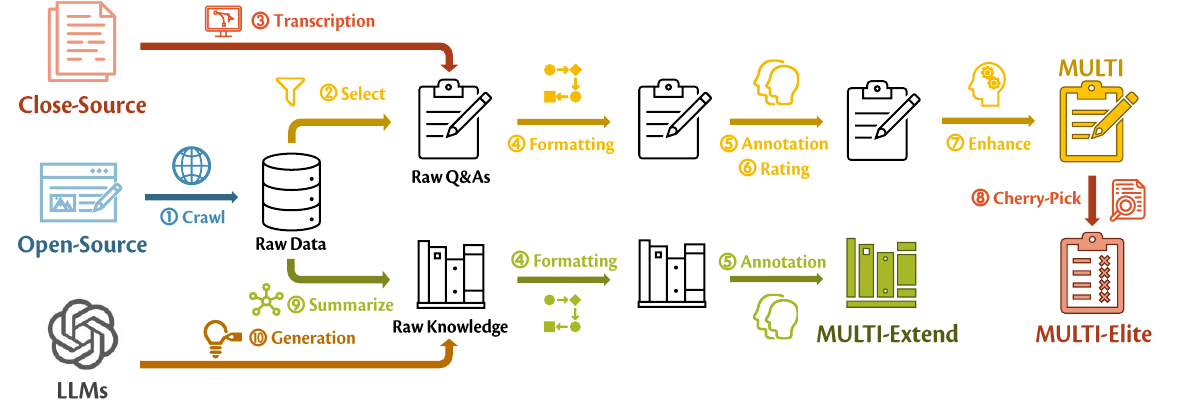}
        \caption{\textbf{The construction pipeline of \Multi. Upper half:} The construction of \multi~and \multihard. (1) Collect large-scale open-source data. (2) Select high-quality and evenly distributed open-source Q\&As. (3) Collect and transcribe questions from closed-source exams. (4) Convert raw data into markdown format using scripts. (5) Skilled annotators refine, split, and label the data. (6) Assess the difficulty and quality of each question. (7) Further refine and increase difficulty through expert review. (8) Select the most challenging items for the \multihard~subset. \textbf{Lower half:} The construction of \multikn. (9) Summarize knowledge from open-source data. (10) Generate additional knowledge via LLMs.}
        \label{figure:pipeline}
    \end{figure}

  The data construction pipeline is shown in \Cref{figure:pipeline}. To develop \multi, we follow several key steps to ensure high-quality and precise annotation. Firstly, we crawl open-source raw question data from the Internet and transcribe closed-source exams from paper documents. Secondly, we format each question and knowledge piece into markdown and \LaTeX~formula format to maintain precision and quality. Thirdly, we revise and refine each question multiple times to prevent data leakage and increase difficulty. Lastly, we rate every question based on its difficulty and content richness.

\paragraph{Data Source}
    We collect more than 2.7M raw data from the Internet, ranging from exams and quizzes from Chinese junior and senior schools and several society exams. We design an algorithm to pick out a proportion of the questions as the fundamental data of our benchmark. The selection is based on the questions' text length, number of images, corresponding subjects, and knowledge pieces, to reach a higher diversity of questions and coverage of knowledge. The details are presented in \ref{sec:algo}. We also collect questions from internal exams and practices of several top universities. After the selection, we obtain over 18K questions as the raw data.

\paragraph{Data Process and Annotation}
The data processing and annotation for our dataset involve a comprehensive series of steps to ensure high-quality, diverse content. 

In the \textbf{Data Pre-process} stage, raw data with formats like HTML, photocopy, hand script, or plain text are refined by removing irrelevant HTML tags, converting text styles into markdown format, and transcribing math functions and chemical structures into \LaTeX~format, with complex tables saved as screenshot images after HTML rendering. OCR tools are utilized for text conversion from photocopies and hand scripts. 

During the \textbf{Data Annotation} stage, an online platform facilitates annotators, primarily skilled undergraduates (involved in the work as authors), in tasks across the format, content, label, and semantic levels. This includes converting content into markdown and \LaTeX, splitting sub-questions into individual ones, evaluating the difficulty and quality, and correcting errors for factual accuracy. 

The \textbf{Data Post-process}  stage employs strategies like formation, disambiguation, distillation, and transformation to enhance question difficulty and diversity, including modifying question formats and reducing assistance information. 

Throughout these stages, we process 2.7 million questions in total and pick out 18,430, incorporating 23,320 scoring points, 7,658 images, and 4,595 knowledge pieces. \multi~highlights a broad diversity in question types, including multiple-choice questions with both single and multiple answers, along with fill-in-the-blank and open-ended writing questions, enriching the testing scenarios.
\footnote{For the sake of simplifying writing, in the following paragraphs we may use abbreviations. We denote multiple-choice questions with a single answer as \textbf{SA} or \textit{Single Answer Choosing} and those with multiple answers as \textbf{MA} or \textit{Multi Answer Choosing}. We use \textbf{FB} for fill-in-the-blank questions and \textbf{OP} for open-ended writing questions.} 
The stages during data processing and annotation significantly increase the diversity and difficulty of the dataset. For details of data processing and annotation, please refer to \ref{sec:app-data}.

\subsection{Statistics}

    \label{sec:stat}

    We provide detailed statistics in \Cref{table:stat}. We also provide the composition of education level and subjects as shown in \Cref{figure:distribution}.

\begin{table}[h]
\begin{minipage}[c]{0.5\textwidth}
\centering
        \footnotesize
        \setlength\tabcolsep{3pt}
            \begin{tabular}{lcc}
                \toprule
                \textbf{Statistics}          & \textbf{Number} & \textbf{Points} \\
                \midrule
                Total Problems               & 17250            & -              \\
                Total Questions              & 18429            & -              \\
                Total Points                 & 22926            & -              \\
                Total Images                 & 7734             & -              \\
                Total Knowledge              & 4595             & -              \\
                \midrule
                Multiple Choices             & 16095(87.33\%)   & 19906(86.83\%) \\
                \quad - Single Answer        & 13949(75.69\%)   & 13952(60.86\%) \\
                \quad - Multiple Answers     & 2143(11.63\%)    & 5954(25.98\%)  \\
                Fill in the blank            & 1439(7.81\%)     & 2223(9.69\%)   \\
                Open-ended Writing           & 797(4.33\%)      & 797(3.48\%)    \\
                Others                       & 98(0.53\%)       & -              \\
                \midrule
                Question with Images         & 7521(40.81\%)    & 8836(38.54\%)  \\
                \quad - Single Image         & 7279(39.50\%)    & 8563(37.35\%)  \\
                \quad - Multiple Images      & 242(1.21\%)      & 273(1.19\%)    \\
                \quad - Image in Choice      & 1179(6.40\%)     & 1181(5.06\%)   \\
                Question with Explanations   & 10565(57.34\%)   & 13186(57.52\%) \\
                Question with Knowledge      & 9048(49.10\%)    & 12919(55.35\%) \\
                \bottomrule
            \end{tabular}
        \caption{\textbf{The statistic overview of \Multi.} Since \multi~contains questions worth multiple points, we provide distribution statistics of both measurements. The statistics are based on \texttt{v1.3.0-20241203}.}
        \label{table:stat}
\end{minipage}
\hspace{0.02\textwidth}
\begin{minipage}[c]{0.41\textwidth}
        \centering
        \includegraphics[width=0.7\textwidth,trim=0 0 0 0,clip]{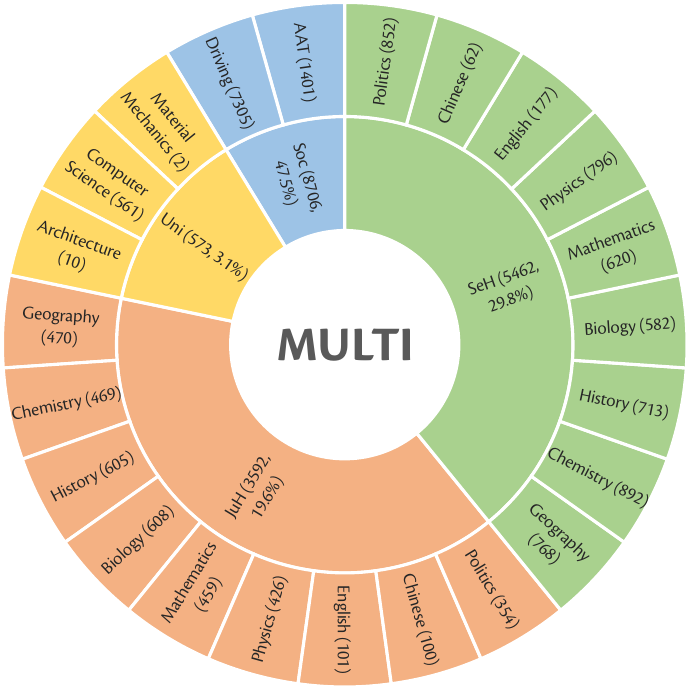}
        \figcaption{\textbf{The question distribution by the education level and subjects.}}
        \label{figure:distribution}

        \vspace{0.5cm}

        \footnotesize
        \centering
        \setlength\tabcolsep{3pt}
        \renewcommand{\arraystretch}{1.0}
        \begin{tabular}{ccccccc}
            \toprule
            \textbf{Type} & \texttt{\#choices} & \texttt{\#A} & \texttt{\#B} & \texttt{\#C} & \texttt{\#D} & \texttt{\#E,F,G...} \\
            \midrule
            \multirow{4.2}{*}{SA}   & 2     & 1813 & 1370 & -    & -    & -   \\
                                    & 3     & 273  & 288  & 265  & -    & -   \\
                                    & 4     & 2192 & 2636 & 2702 & 2379 & -   \\
                                    & 5     & 3    & 6    & 4    & 4    & 0   \\
            \hdashline
            MA   & 3-13                     & 1466 & 1571 & 1516 & 1307 & 92  \\
            \midrule
            Total &2-13                     & 5747 & 5871 & 4490 & 3694 & 92  \\ 
            \bottomrule
        \end{tabular}
        \tabcaption{\textbf{The answer distribution for multiple-choice questions.}}
        \label{table:data_dist}
\end{minipage}
\end{table}

    Our benchmark showcases a remarkable diversity in the choice setting of multiple-choice questions, encompassing options that range from 2 to as many as 13. Furthermore, it includes questions that vary in the number of correct answers, from questions with a unique correct option to those with multiple correct choices. We provide the distribution of choices in multiple-choice questions as shown in \Cref{table:data_dist}. Each row corresponds to a different total number of options available in the questions. The columns represent the frequency of each specific choice option. The table showcases a well-balanced distribution of choices. Notably, the distribution reveals a higher frequency of questions with four choices and a single correct answer, indicating a common format in multiple-choice questions. 


    In addition to multiple-choice questions, our benchmark also includes a substantial number of fill-in-the-blank and open-ended questions, creating a diverse and comprehensive range of testing scenarios. Moreover, we have incorporated unique open-response questions that require creative answers, such as drawings. It is important to note that these open-response questions are not included in our formal evaluation and scoring procedures, for they are primarily proposed for qualitative research and development in the field of MLLMs. Our benchmark is carefully designed to thoroughly assess and enhance the ability of MLLMs to process and respond to various question types, resembling real-world scenarios.

    \subsection{The \Multihard~Set}
\label{sec:3.3}
   We have selected 500 objective questions to create a hard subset. These multiple-choice and fill-in-the-blank questions are more evenly distributed across various educational levels and subjects and have been evaluated by annotators as high in difficulty and quality, featuring detailed content and complex reasoning requirements. In the selection process, the performance of several early-released models during earlier evaluation is also taken into consideration. We especially focused on poorly answered questions by GPT-4V\cite{gpt4v}, the state-of-the-art model at the time. As a result, these models may get lower scores compared to the random baseline in Table \ref{table:extreme_result}. Since a complete run on \multi~could be costly for some models, we encourage researchers to run their experiments on these 500 high-quality questions. 

    \subsection{The \Multikn~Background Knowledge Dataset}

\label{sec:3.4}
    External knowledge is crucial to provide critical information that assists in solving questions using the In-Context Learning (ICL) abilities. Some of the raw questions retrieved from the Internet have corresponding knowledge pieces attached. We also collect more knowledge pieces for uncovered questions with the assistance of LLMs and external knowledge sources (e.g., New Bing\footnote{\url{https://bing.com/new}} and Wikipedia\footnote{\url{https://wikipedia.org}}). We conduct annotations on these knowledge pieces to confirm the correctness of the content and present them in the \multikn~dataset. This dataset consists of about 4.6K knowledge pieces, designed to test the in-context learning abilities and knowledge transfer skills of models. This dataset provides comprehensive insights into the capabilities and limitations of multimodal LLMs, opening new pathways for research exploration.

\subsection{Comparison with Existing Benchmarks}
\label{sec:3.5}

\begin{table*}[h]
    \centering
    \footnotesize
    \setlength\tabcolsep{4pt}
    \renewcommand{\arraystretch}{1.0}
    \begin{tabular}{lcccccccccccccc}
        \toprule
        \multirow{2.5}{*}{\textbf{Benchmark}} & \multirow{2.5}{*}{\textbf{Lang}} & \multicolumn{5}{c}{\textbf{Size}} & \multicolumn{3}{c}{\textbf{Image}}  & \multicolumn{4}{c}{\textbf{Answer Type}}  & \multirow{2.5}{*}{\textbf{Source}}\\
        \cmidrule(rl){3-7}\cmidrule(rl){8-10}\cmidrule(rl){11-14}
        ~                                &        & \textbf{Sub} & \textbf{Q}    & \textbf{Ana}   & \textbf{Img}     & \textbf{Kn}  & \textbf{NI} & \textbf{SI}           & \textbf{MI}           & \textbf{SA}           & \textbf{MA}           & \textbf{FB}           & \textbf{OP}           &                                \\
        \midrule
        VQA~\cite{antol2015vqa}          & en     & 36  & 764k & -     & 265k    & -  & \; &\V   &\; &\; &\;   &\V &\; & Repurposed \\
        ScienceQA~\cite{lu2022learn}     & en     & 21  & 21k  & 19k   & 10k     & 0.3k&\V &\V   &\; &\V &\; &\; &\; & Textbooks \\
        SciBench~\cite{wang2023scibench} & en     & 6   & 0.8k & -     & 0.1k    & -  & \; &\V &\; &\; &\; &\V   &\V   & Textbooks \\
        M3Exam~\cite{zhang2023m3exam}    & 9 langs & 4    & 12k   & -   & 3.1k & -    & \V &\V   &\; &\V &\; &\; &\;   & Exams \\
        AGIEval~\cite{zhong2023agieval}  & zh, en & 20  & 8k   & a few & -       & -  & \V &\;    &\; &\V   &\V   &\V   &\; & Exams \\
        CMMU~\cite{ijcai2024p92}         & zh     & 7  & 4k   &  -  & 4k    & - & \;  &  \V  & \;   &   \V  & \V   &   \V &   \;   & Exams \\
        GAOKAO-MM~\cite{zong2024gaokao}  & zh     & 8  &  0.6k  & -  & 0.9k & -   & \;  & \V   & \V   &   \V  & \;   &  \;  &  \;   & Exams \\
        MMBench~\cite{liu2023mmbench}    & zh/en     & 20  & 3k   & -     & 3k      & -  & \; &\V   &\; &\V   &\; &\; &\; & Web, repurposed \\
        SEED-Bench~\cite{li2023seed}     & en     & 12  & 19k  & -     & 19k+    & -  & \; &\V   &\V   &\V   &\; &\; &\; & Anno. \\
        SEED-Bench-2~\cite{li2023seed2}  & en     & 27  & 24k  & -     & 22k+    & -  & \; &\V   &\V   &\V   &\; &\; &\; & Anno. \\
        MLLM-Bench~\cite{ge2023mllm}     & en     & 42  & 0.4k & -     &  0.4k   & -  & \; &\V   &\;   &\;   &\; &\; &\V & Anno. \\
        Touchstone~\cite{bai2023touchstone} & en & 27  & 0.9k & -     & 0.9k    & -  & \; &\V   &\V   &\;   &\; &\; &\V & Anno. \\
        AlignMMBench~\cite{wu2024alignmmbench} & zh     & 13  & 5k  & -    & 1k        & -  & \; &\V   &\;    &\;    &\;  &\; &\V & Web, generated \\
        C-Eval~\cite{huang2023ceval}     & zh     & 52  & 14k  & a few & -       & -  & \; &\; &\; &\V   &\; &\; &\; & Exams, web \\
        SciEval~\cite{sun2023scieval}    & en     & 3   & 18k  & -     & -       & -  & \V &\; &\; &\V   &\; &\; &\; & Web, repurposed \\
        MMMU~\cite{yue2023mmmu}          & en     & 30  & 12k  & 2k    & 11k+    & -  & \; &\V   &\V   &\V   &\; &\V &\; & Anno., web, textbooks \\
        CMMMU~\cite{zhang2024cmmmu}      & zh     & 30  & 12k  & -    & 12k    & -  & \; &\V   &\V   &\V   &\; &\V &\; & Anno., web, textbooks \\
        \midrule
        \multi~(ours)                    & zh     & 23  & 18k  & 10k+  & 7.7k   & 4.6k & \V &\V   &\V   &\V   &\V   &\V   &\V   & Anno., exams, web  \\
        \bottomrule
    \end{tabular}
    \caption{\textbf{The comparison between \Multi~and other existing knowledge-based benchmarks.} Sub: Subject or Field, Q: Question, Ana: Analysis or Explanations, Img: Images, Kn: Knowledge or Lecture. NI: the question with pure text, SI: the question with a single image, MI: the question with multiple images. SA: multiple-choice question with single correct answer, MA: multiple-choice question with multiple correct answers, FB: fill-in-the-blank question (no more than 10 words), OP: open-ended writing question (more than 10 words). Anno.: Annotation.}
    \label{table:benchmarks}
\end{table*}
    
    \multi~demonstrates a comprehensive blend of features that surpasses existing benchmarks in several dimensions. Notably, \multi~covers a wide array of subjects and a substantial number of questions (18K), as well as over 10K analysis and 4.6K extensive knowledge content, which is considerably larger than most benchmarks, ensuring a broad and diverse testing environment. \multi~possesses 7.7K images, which is essential for benchmarking MLLMs that require visual understanding alongside textual information. The inclusion of both single and multiple image questions, as well as a variety of answer types, makes \multi~a versatile and challenging benchmark. Furthermore, the questions without images also test the MLLMs' ability to deal with plain text information. Meanwhile, the various sources, complex annotation, and processing stages provide sufficient augmentation to alleviate data leakage. \multi~not only encompasses variations of classic questions but also includes recently updated questions, which significantly enhances its diversity.

    We list the features of existing benchmarks and make a comparison with \multi~in \Cref{table:benchmarks}. We believe that \multi~assembles the most advantages of the existing benchmarks and is sure to provide a good option for the community to test the capabilities of their MLLMs.

    \section{Experiments}
    \label{sec:exp} In this section, we conduct a comprehensive experiment and several ablation studies. We firstly introduce baselines, settings, and metrics in \Cref{sec:4.1}, \Cref{sec:4.2}, and \Cref{sec:metrics}. The main experiment analysis is in \Cref{sec:main}, followed by several research questions and ablation study results from \Cref{sec:hard} to \Cref{sec:error}. We conclude our findings and thoughts as take-home messages in \Cref{sec:takehome}.
    
    \subsection{Baselines}
    \label{sec:4.1}
    \paragraph{Large Models.}  We evaluate a wide range of leading-edge MLLMs that support Chinese, including both open-sourced models \cite{cllava,bai2023qwen,wang2024qwen2,viscpm,du2022glm,chen2023internvl,yi} and closed-source models \cite{gpt4omini,gpt4v,geminiteam2023gemini,claude3,gpt4o}. We evaluate these models with both multimodal input and text-only input to verify the information gained from input images. We also select several most capable and representative LLMs \cite{chen2022dialogzoo,sun2023moss,chatgpt,gpt4,qwen2,zhang2023internlm,cai2024internlm2} for comparison. We also add ablation studies to verify the improvement by giving the OCR information or captions to these LLMs. For detailed information about all models evaluated in~\multi, please refer to \Cref{table:models} in \ref{sec:model}.

    \paragraph{Human Experts.} Seven human experts from top universities in China are involved in the human evaluation. The complete human evaluation set includes all 500 questions from \multihard~and another 500 randomly sampled questions from \multi. Each question was assigned to two experts based on the subjects they had chosen in the Gaokao. We follow the rule in MMMU\cite{yue2023mmmu} where the experts are allowed to consult their textbooks and \multikn~but are forbidden to search the Internet. We calculate the score of \multi~as the weighted average of each subject.
    
    \paragraph{Random Baseline.} We also provide a random baseline and a most-frequent baseline. For the random baseline, we randomly choose one choice in SA, randomly select/drop each choice in MA, and randomly choose from \texttt{[A, B, C, D, 1, 2, 3, 4, True, False]} in OP, and simply take the question content as the answer.

    \begin{wrapfigure}[17]{r}{0.45\textwidth}
        \centering
        \vspace{-1.2cm}
        \includegraphics[width=0.45\textwidth,trim=10 300 10 10,clip]{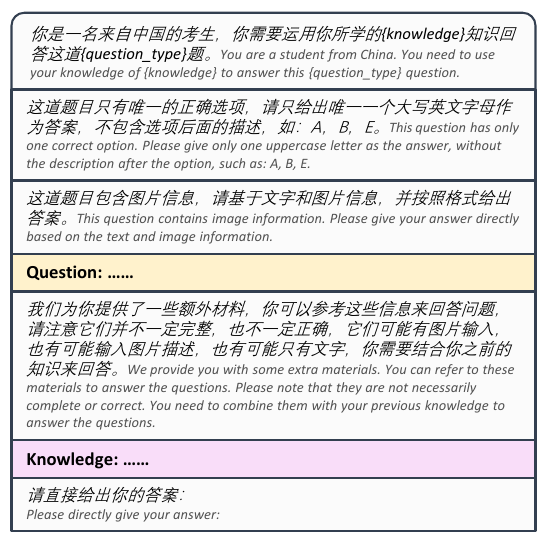}
        \caption{\textbf{An example of the prompt template.} It is used when evaluating a multiple-choice question with image context, knowledge piece, and single correct answer.}
        \label{figure:prompts}
    \end{wrapfigure}

    \subsection{Settings}
    \label{sec:4.2}

    \paragraph{Prompt} We use specialized prompts for each question, an example shown in \Cref{figure:prompts}. The prompts are designed carefully according to the features of each type of question and the answer patterns expected. We also modify the input format to fit into official inference guidelines. The complete collection of prompts is presented in \ref{sec:prompts}.

    \paragraph{Image} \multi~includes questions with either none, single, or multiple images. Most MLLMs accept text accompanied by one image as input or a pure-text input. For questions with a single image, the image and text are fed in one turn. We simply drop image information when evaluating LLMs. 
    
    For pure-text questions, we use the text as input. For some models like VisCPM, InternVL-1.1, and Yi-VL-34B, which compulsorily demand an image in each turn, we feed the model a blank image with color set to RGB(0,0,0) along with plain text in evaluation. 
    
    For questions with multiple images, as the positions of images matter a lot, e.g., a multiple-choice question where each choice is an image, special patterns with \texttt{[IMAGE\_\{index\}]} are used to indicate the position and order of images. Qwen-VL, GPT-4V, and Gemini Vision naturally support multiple images as input in one turn, while VisCPM and VisualGLM support only one image as input in one turn. We adopt the strategy of splitting the content into multiple segments divided by each image and feeding them into the MLLM sequentially as rounds of conversation, where the MLLM receives each segment along with the corresponding image. We tune our prompts so that the MLLM may receive all the information but should only give a finalized answer after we show a signal that the question ends. The prompt we use in multi-turn input is shown in \Cref{figure:prompts_all}. As the released versions of Chinese-LLaVA, InternVL-1.1, and Yi-VL-34B do not support multiple images as input, currently only the first image is used for evaluating each question.

    \subsection{Metrics}
    \label{sec:metrics}

    We focus on objective questions with a certain answer, including multiple-choice and fill-in-the-blank questions. We also give a score to each subjective open-ended question based on the similarity to the reference answer. We report the score ratio of all questions as the final metric. The metrics we use for each type of question:

    \paragraph{Multiple-choice with Single Answer (SA)} Each question worth one point. We calculate the accuracy of the given answer. 

    \paragraph{Multiple-choice with Multiple Answers (MA)} We follow the metric used in Gaokao and define the total points of an MA question as the number of correct choices, and each correct choice selected is rewarded one point. If the given answer contains any wrong choices, the score will be counted as zero. We report the score ratio (\texttt{\# points / \# total points}) as the metric. We also report accuracy as a more rigorous metric, where only correctly giving all the choices without wrong ones will be granted points. For example, a question with a correct answer \texttt{ACE} worth 3 points, and the incomplete answer \texttt{AC} will be granted 2 points, and the answer \texttt{BC} or \texttt{ABCE} will be granted 0 points. However, on calculating accuracy, none will be counted, except for \texttt{ACE}, which will be calculated as correct.

    \paragraph{Fill in the Blank (FB)} We define the total points of a blank-filling question as the number of the blanks marked as \texttt{[MASK]}. It is required in the prompts that each line of the given answer corresponds to a blank in order. We follow the strictest standard of \textit{exact match}. Therefore, only answers exactly matching one of the standard answers will be granted points.

    \paragraph{Open-ended Question (OP)} The points and counting method is similar to FB, but we use a loose standard and report the normalized ROUGE-L~\cite{lin-2004-rouge} score for each point. Please note that since the reference answer may be concise or in detail, and there could be other possible answers, the lower score ratio of OP questions is expected and reasonable.

    \subsection{Main Experiment Results}
    \label{sec:main}

    In this sub-section we report the performance on \multi~as detailed in Table \ref{table:overall_result}, \ref{table:image_result}, and \ref{table:type_result} along with analysis below:

    \begin{wraptable}[22]{r}{0.6\textwidth}
            \footnotesize
        \centering
        \vspace{-0.8cm}
        \setlength\tabcolsep{5pt}
        \renewcommand{\arraystretch}{1.0}
        \begin{tabular}{lcccccc}
            \toprule
            \textbf{Model}    & \cellcolor{rmbblue!50}\textbf{Overall} & \textbf{JuH} & \textbf{SeH} & \textbf{Uni} & \textbf{Driv} & \textbf{AAT} \\
            \midrule
            Random            & \cellcolor{rmbblue!50}24.8             & 14.0         & 17.2         & 20.4         & 35.8          & 24.6         \\
            Most              & \cellcolor{rmbblue!50}28.4             & 14.2         & 19.7         & 20.5         & 42.8          & 23.3         \\
            Expert            & \cellcolor{rmbblue!50}86.1             & 87.8         & 82.9         & 88.2         & 91.7          & 76.7         \\
            \midrule
            VisualGLM         & \cellcolor{rmbblue!50}31.1             & 22.2         & 25.6         & 23.6         & 40.9          & 24.9         \\
            VisCPM            & \cellcolor{rmbblue!50}33.4             & 25.2         & 28.1         & 23.0         & 43.4          & 23.7         \\
            Chinese-LLaVA     & \cellcolor{rmbblue!50}28.5             & 21.1         & 25.4         & 20.7         & 35.8          & 21.8         \\
            Qwen-VL           & \cellcolor{rmbblue!50}39.0             & 32.6         & 32.9         & 27.2         & 49.3          & 26.4         \\
            Yi-VL-34B         & \cellcolor{rmbblue!50}55.3             & 46.6         & 46.0         & 45.4         & 71.1          & 26.5         \\
            InternVL-1.1      & \cellcolor{rmbblue!50}44.9             & 39.3         & 36.5         & 30.6         & 57.7          & 24.8         \\
            InternVL2-8B      & \cellcolor{rmbblue!50}67.9             & \U{63.7}     & \U{58.9}     & \U{59.6}     & 83.8          & 26.0         \\
            MiniCPM-V-2.6     & \cellcolor{rmbblue!50}55.1             & 46.3         & 42.8         & 38.0         & 73.7          & 27.4         \\
            Qwen2-VL-7B       & \cellcolor{rmbblue!50}\U{68.6}         & 60.4         & 57.8         & 50.8         & \U{87.5}      & \U{28.7}     \\
            Qwen2-VL-72B      & \cellcolor{rmbblue!50}\B{76.9}         & \B{66.3}     & \B{69.5}     & \B{69.9}     & \R{\B{94.3}}  & \B{34.0}     \\
            \midrule
            GPT-4V            & \cellcolor{rmbblue!50}63.7             & 58.5         & 52.9         & 59.0         & 80.1          & 26.2         \\
            Gemini Vision Pro & \cellcolor{rmbblue!50}53.7             & 48.2         & 45.2         & 41.7         & 67.4          & 27.0         \\
            GPT-4o-mini       & \cellcolor{rmbblue!50}58.2             & 51.7         & 43.9         & 49.6         & 77.1          & 25.8         \\
            GPT-4o (0806)     & \cellcolor{rmbblue!50}69.9             & 65.6         & 58.6         & \U{62.2}     & \B{87.1}      & 28.0         \\
            Gemini 1.5 Flash  & \cellcolor{rmbblue!50}64.9             & 63.2         & 56.0         & 52.6         & 78.7          & 26.7         \\
            Claude 3.5 Sonnet & \cellcolor{rmbblue!50}57.2             & 38.7         & 42.3         & 26.1         & 83.3          & \U{28.9}     \\
            GPT-4o (1120)     & \cellcolor{rmbblue!50}\U{70.0}         & \U{65.9}     & \U{59.0}     & 60.2         & \U{86.9}      & 28.2         \\
            Gemini 1.5 Pro    & \cellcolor{rmbblue!50}\B{71.8}         & \B{71.6}     & \B{64.3}     & \B{65.3}     & 84.2          & \B{29.3}     \\
            \bottomrule
        \end{tabular}
        \caption{\textbf{The main performance (\%) of all models evaluated on \Multi.} JuH: Level of Junior High School, SeH: Level of Senior High School, Uni: Level of University, Driv: Chinese Driving Test, AAT: Administrative Aptitude Test.  We categorized the models into two groups based on whether they are accessed through model weights or via an API service. Models in each group are sorted by their release date. The highest scores in each group are \B{in bold}, while the second highest scores are \U{underlined}. The scores higher than the human expert baseline are \R{highlighted}.}
        \label{table:overall_result}
    \end{wraptable}

    \paragraph{Most models are struggled with \Multi.}
    \label{para:overall}
    By the beginning of 2024, GPT-4V achieves a score of 63.7\%, while by the end of 2024, the most powerful competitor, Qwen2-VL-72B, achieves a score of 76.9\%, underscoring \multi's complexity and challenge even after near a year of its first release. But more importantly, we observe a huge improvement in performance for open-sourced models like QwenVL and InternVL series, indicating the gap between open-source MLLMs and closed ones is becoming increasingly narrow. However, the gap between these models and human experts remains significant. 

\paragraph{Analysis from the perspective of education level and subjects:}
\label{para:subject}
In \Cref{table:overall_result}, we also present the performance categorized by educational levels and subjects. The performance trends for non-society level questions remain consistent with the overall results observed, while the performance of these three levels is similar.

For questions at the society level, i.e., Driv. and AAT, we anticipate higher scores on the Driving Test. This may be caused by a larger percentage of judgmental questions (in the format of SA with two options), as well as its nature with knowledge of regulations. We are happy to say that Qwen2-VL-72B has surpassed the human expert baseline in this subject.

Furthermore, questions from the Administrative Aptitude Test (AAT), which typically include at least one image and often examine skills in image pattern recognition (illustrated in the first two examples in the left column of \Cref{figure:examples}), tend to have scores around or below the random baseline. Even the strongest competitor, Qwen2-VL-72B, shows limited success, with a performance of only 34.0\% on these questions. This underscores the significant challenge posed by multimodal questions.

\paragraph{Analysis from the perspective of different numbers of images:} 
\label{para:image}
 In \Cref{table:image_result}, we present the performance categorized by image number. While the highest score on the None-image set achieves 86.9\% by Qwen2-VL-72B, a significant drop in performance is observed when answering questions with more than one image. Only Gemini 1.5 Pro manages to get a 50.0\% score while other strong models are around 40-41\%. It is evident that questions requiring more images are more challenging. The high performance on the NI set may also be caused by the large portion of easy-driving questions as mentioned before.
 

\paragraph{Analysis from the perspective of different question types.} 
\label{para:type}
In \Cref{table:type_result}, we present the performance categorized by question type. A majority of the models achieve their highest scores on the Single Answer Choosing (SA) set, with lower performance on the Multiple Answers Choosing (MA) set. A notable discrepancy is observed between the scores for the MA set and its accuracy, highlighting the smaller models' inability to identify all correct options accurately. While Qwen2-VL-72B once again beat the human expert baseline.

For the Fill-in-the-Blank (FB) set, which requires short but accurate matches, the scores further decline. This is partially due to failure to follow the specified instructions, often leading to correct responses being presented in an unacceptable format. Gemini 1.5 and GPT-4o models are more capable in FB questions and surpass Qwen2-VL-72B by large.  

Furthermore, we note significantly lower scores on the Open-ended Writing (OP) set in comparison to the FB set. None of the models gets even half of the 46.4 \% human expert baseline on the OP set. This also suggests that \multi~minimizes the risk of data leakage during its construction and poses considerable challenges for models in generation across modalities.

    \begin{table}[h]

        \begin{minipage}[c]{0.48\textwidth}
            \footnotesize
            \centering
            \setlength\tabcolsep{5pt}
            \renewcommand{\arraystretch}{1.0}
            \begin{tabular}{lccc}
                \toprule
                \textbf{Model}    & \textbf{NI}  & \textbf{SI} & \textbf{MI} \\
                \midrule
                Random            & 27.2         & 21.0        & 16.9        \\
                Most              & 32.5         & 22.0        & 17.7        \\
                Expert            & 86.6         & 85.3        & 81.8        \\
                \midrule
                VisualGLM         & 35.1         & 25.2        & 9.7         \\
                VisCPM            & 36.8         & 28.4        & 16.6        \\
                Chinese-LLaVA     & 32.3         & 22.6        & 17.8        \\
                Qwen-VL           & 43.2         & 32.7        & 20.7        \\
                Yi-VL-34B         & 63.8         & 42.0        & 24.5        \\
                InternVL-1.1      & 50.9         & 35.5        & 25.1        \\
                InternVL2-8B      & \U{78.7}     & 51.6        & 28.7        \\
                MiniCPM-V-2.6     & 63.8         & 41.8        & 24.0        \\
                Qwen2-VL-7B       & 78.1         & \U{53.7}    & \B{34.2}    \\
                Qwen2-VL-72B      & \R{\B{86.9}} & \B{61.5}    & \B{41.1}    \\
                \midrule
                GPT-4V            & 74.5         & 46.9        & 28.1        \\
                Gemini Vision Pro & 62.5         & 40.0        & 24.5        \\
                GPT-4o-mini       & 67.7         & 43.0        & 30.4        \\
                GPT-4o (0806)     & \U{80.0}     & 54.1        & \U{40.9}    \\
                Gemini 1.5 Flash  & 73.8         & 50.9        & 40.0        \\
                Claude 3.5 Sonnet & 68.3         & 39.7        & 23.1        \\
                GPT-4o (1120)     & 79.6         & \U{54.9}    & \U{40.9}    \\
                Gemini 1.5 Pro    & \B{80.2}     & \B{58.5}    & \B{50.0}    \\
                \bottomrule
            \end{tabular}
            \caption{\textbf{Performance (\%) on each image type of \Multi.}}
            \label{table:image_result}
        \end{minipage}
        \hspace{0.04\textwidth}
        \begin{minipage}[c]{0.48\textwidth}
            \footnotesize
            \centering
            \setlength\tabcolsep{5pt}
            \renewcommand{\arraystretch}{1.0}
            \begin{tabular}{lccccc}
                \toprule
                \B{Model}         & \B{SA}   & \B{MA}       &   \makecell{\B{MA} \\\B{Acc.}}             & \B{FB}       & \B{OP}   \\
                \midrule
                Random            & 30.8     & 22.5         & 6.5          & 0.9          & 3.4      \\
                Most              & 30.7     & 36.3         & 21.0         & 1.6          & 3.4      \\
                Expert            & 91.4     & 81.1         & 70.3         & 83.4         & 46.4     \\
                \midrule
                VisualGLM         & 37.9     & 30.2         & 1.9          & 0.7          & 3.6      \\
                VisCPM            & 41.7     & 27.7         & 0.0          & 3.8          & 14.1     \\
                Chinese-LLaVA     & 34.5     & 26.9         & 3.9          & 2.4          & 8.4      \\
                Qwen-VL           & 49.8     & 29.4         & 2.8          & 5.8          & 13.7     \\
                Yi-VL-34B         & 61.3     & 42.0         & 36.4         & 14.6         & 8.9      \\
                InternVL-1.1      & 56.4     & 33.4         & 2.1          & 14.2         & 13.1     \\
                InternVL2-8B      & 72.6     & \U{74.2}     & \U{64.0}     & \B{40.9}     & 17.5     \\
                MiniCPM-V-2.6     & 64.7     & 53.2         & 31.3         & 15.2         & 12.3     \\
                Qwen2-VL-7B       & \U{75.0} & 72.7         & 62.1         & 35.0         & \U{18.5} \\
                Qwen2-VL-72B      & \B{83.5} & \R{\B{84.7}} & \R{\B{78.1}} & \U{35.1}     & \B{19.5} \\
                \midrule
                GPT-4V            & 67.1     & 70.6         & 58.2         & 42.4         & 11.7     \\
                Gemini Vision Pro & 59.4     & 54.4         & 24.3         & 30.5         & 12.5     \\
                GPT-4o-mini       & 65.8     & 56.6         & 46.3         & 30.3         & 12.4     \\
                GPT-4o (0806)     & \B{74.5} & \U{74.6}     & \U{63.1}     & 47.6         & \U{17.2} \\
                Gemini 1.5 Flash  & 68.4     & 68.5         & 52.1         & \U{50.9}     & 14.1     \\
                Claude 3.5 Sonnet & 63.0     & 64.9         & 52.9         & 16.6         & 9.4      \\
                GPT-4o (1120)     & \U{74.4} & 74.2         & 63.0         & 49.4         & \B{17.7} \\
                Gemini 1.5 Pro    & 74.3     & \B{76.8}     & \B{64.9}     & \B{62.6}     & 14.6     \\
                \bottomrule
            \end{tabular}
            \caption{\textbf{Performance (\%) on each type of questions of \Multi.} MA Acc.: Accuracy of MA questions.}
            \label{table:type_result}
        \end{minipage}

    \end{table}

    \vspace{-1em}
We keep evaluating new models and updating our leaderboard on our \href{https://opendfm.github.io/MULTI-Benchmark/\#leaderboard}{\U{official website}} \footnote{\url{https://opendfm.github.io/MULTI-Benchmark/\#leaderboard}}. Readers can find other useful links in~\ref{sec:app-links}.

\subsection{Is \Multihard~more challenging?}

\label{sec:hard}

We conduct evaluations on \multihard, as outlined in \Cref{table:extreme_result}, which includes 500 specifically chosen questions. These questions are selected based on pre-annotated quality and difficulty scores, in addition to the evaluation results on \multi~discussed in \Cref{sec:main}. The selection aims to ensure a more balanced distribution but also picks those high-quality questions that bring a challenge to strong MLLMs. 

Qwen2-VL-72B and Gemini 1.5 Pro achieve the highest score on \multihard~with 53.1\% and 46.9\%, while scores for other models vary between 10.5\% and 40.0\%. This highlights the substantial challenge presented by \multihard, indicating significant potential for improvement in tackling extremely difficult questions that require in-depth image understanding and intricate reasoning across modalities. 

It is important to highlight the accuracy of multiple answers choosing (MA Acc.) as the most demanding task for MLLMs, necessitating a thorough grasp of the relationships between the choices and the questions, and reflecting model's reliability of selecting all answers correctly. Meanwhile, the FB set has the largest gap between the highest model performance and human expert baseline, indicating a great challenge for models to answer accurately in these questions.


    \begin{table}[h]
        \begin{minipage}[c]{0.55\textwidth}
            \footnotesize
            \centering
            \setlength\tabcolsep{2.1pt}
            \renewcommand{\arraystretch}{1.0}
            \begin{tabular}{lcccccccc}
                \toprule
                \textbf{Model} & \cellcolor{rmbblue!50}\textbf{Overall} & \textbf{SA} & \textbf{MA} & \makecell{\textbf{MA} \\\textbf{Acc.}} & \textbf{FB} & \textbf{NI} & \textbf{SI} & \textbf{MI}  \\
                \midrule
                Random            & \cellcolor{rmbblue!50}17.7             & 28.0        & 13.3        & -           & 0.5         & 14.6        & 19.7        & 16.0        \\
                Most              & \cellcolor{rmbblue!50}24.5             & 26.4        & 33.2        & 16.3        & 1.6         & 22.2        & 26.4        & 20.3        \\
                Expert            & \cellcolor{rmbblue!50}73.1             & 78.4        & 67.8        & 54.8        & 70.8        & 69.5        & 76.2        & 66.1        \\
                \midrule
                VisualGLM$^\dagger$         & \cellcolor{rmbblue!50}12.8             & 14.5        & 16.6        & 0.0         & 0.8         & 16.2        & 11.7        & 6.8         \\
                VisCPM$^\dagger$            & \cellcolor{rmbblue!50}13.0             & 10.4        & 22.0        & 0.0         & 0.8         & 10.3        & 14.2        & 15.3        \\
                Chinese-LLaVA$^\dagger$     & \cellcolor{rmbblue!50}12.3             & 15.7        & 13.1        & 1.0         & 1.6         & 13.7        & 11.0        & 15.3        \\
                Qwen-VL$^\dagger$           & \cellcolor{rmbblue!50}10.5             & 7.2         & 19.3        & 1.9         & 0.8         & 8.5         & 10.8        & 16.9        \\
                Yi-VL-34B$^\dagger$         & \cellcolor{rmbblue!50}26.2             & 33.0        & 29.0        & 8.7         & 3.2         & 32.5        & 22.7        & 25.4        \\
                InternVL-1.1$^\dagger$      & \cellcolor{rmbblue!50}20.7             & 24.8        & 23.2        & 0.0         & 4.8         & 17.9        & 21.0        & 28.8        \\
                InternVL2-8B      & \cellcolor{rmbblue!50}36.3             & 38.5        & 39.5        & 27.6        & \U{23.3}    & \U{40.4}    & \U{39.5}    & 23.3        \\
                MiniCPM-V-2.6     & \cellcolor{rmbblue!50}21.3             & 23.3        & 26.1        & 7.6         & 5.8         & 18.4        & 22.6        & 23.4        \\
                Qwen2-VL-7B       & \cellcolor{rmbblue!50}\U{40.0}         & \U{43.5}    & \U{43.3}    & \U{32.4}    & \U{23.3}    & 39.9        & 38.4        & \B{50.0}    \\
                Qwen2-VL-72B      & \cellcolor{rmbblue!50}\B{53.1}         & \B{55.3}    & \B{62.8}    & \B{54.3}    & \B{25.8}    & \B{63.6}    & \B{49.6}    & \U{37.5}    \\
                \midrule
                GPT-4V$^\dagger$            & \cellcolor{rmbblue!50}14.0             & 5.3         & 25.5        & 15.4        & 12.0        & 7.3         & 14.9        & 33.9        \\
                Gemini Vision Pro$^\dagger$ & \cellcolor{rmbblue!50}12.4             & 5.3         & 21.2        & 5.8         & 12.0        & 6.8         & 12.0        & 37.3        \\
                GPT-4o-mini       & \cellcolor{rmbblue!50}18.2             & 22.3        & 18.5        & 11.5        & 7.2         & 17.1        & 17.4        & 28.8        \\
                GPT-4o (0806)     & \cellcolor{rmbblue!50}36.0             & \U{43.1}    & 37.5        & 24.0        & 15.2        & 34.2        & 35.9        & 44.1        \\
                Gemini 1.5 Flash  & \cellcolor{rmbblue!50}33.4             & 32.3        & 38.3        & 21.9        & \U{25.8}    & 33.8        & 30.7        & 50.0        \\
                Claude 3.5 Sonnet & \cellcolor{rmbblue!50}34.3             & 38.1        & 38.6        & \U{30.8}    & 16.0        & 33.3        & 33.3        & 45.8        \\
                GPT-4o (1120)     & \cellcolor{rmbblue!50}\U{36.3}         & 42.2        & \U{42.1}    & 25.7        & 17.5        & \U{36.0}    & \U{36.3}    & \U{56.2}    \\
                Gemini 1.5 Pro    & \cellcolor{rmbblue!50}\B{46.9}         & \B{47.8}    & \B{47.9}    & \B{33.3}    & \B{42.5}    & \B{43.4}    & \B{46.0}    & \B{65.6}    \\
                \bottomrule
            \end{tabular}
            \tabcaption{\textbf{The performance (\%) of all models evaluated on \Multihard.} $\dagger$: these models are chosen to guide the \multihard~selection process, thus the performance of these models is relatively lower.}
            \label{table:extreme_result}
        \end{minipage}
        \hspace{0.04\textwidth}
    \begin{minipage}[c]{0.40\textwidth}
            \includegraphics[width=1.0\linewidth,trim=0 0 0 0,clip]{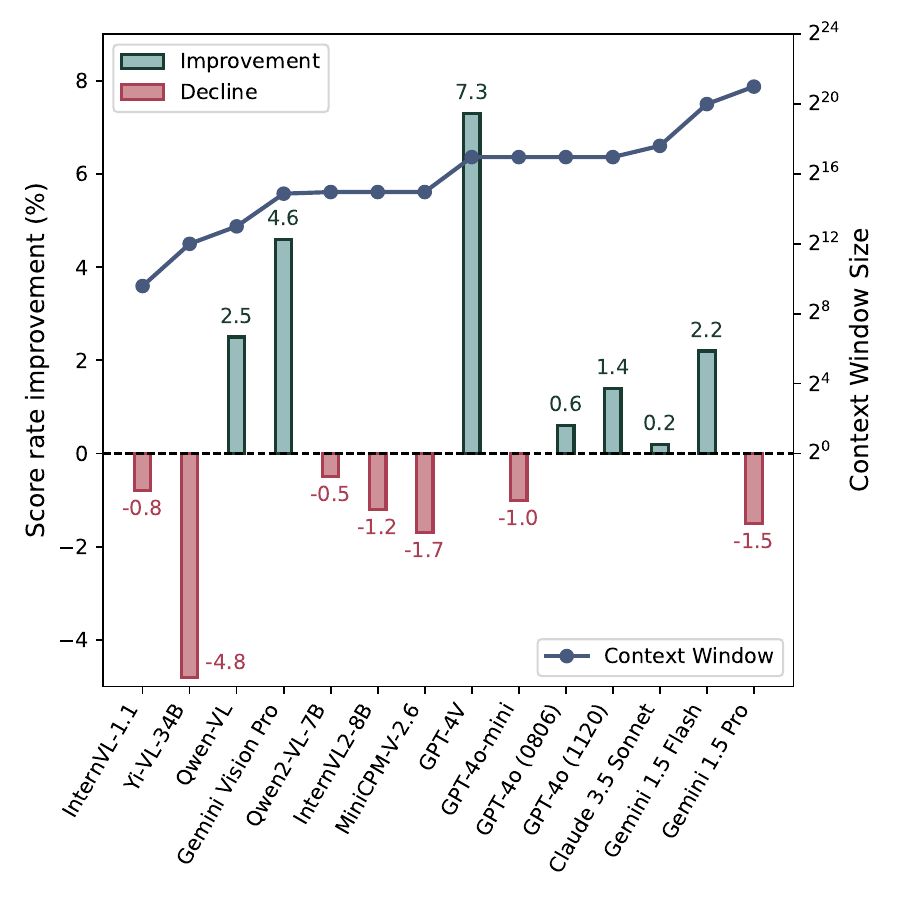}
            \figcaption{\textbf{Performance improvement with \Multikn~on \Multihard.} The context window size of each model is plotted in log scale.}
            \label{fig:kn_result}
        \end{minipage}
    \end{table}

    \vspace{-1.5em}

\subsection{Can model utilize \Multikn?}
\label{sec:kn}

The significant challenges posed by \multihard~prompt further investigation into the In-Context Learning (ICL) capabilities of MLLMs through the utilization of the \multikn~knowledge set. This set is designed to include relevant concepts and frequently utilized solutions related to the problems. Notably, the average number of tokens per question escalates from 65 to 250, and further to 850, following the integration of prompts and the adoption of \multikn, with the most extensive examples surpassing 16,000 tokens. 

The study is conducted on several selected MLLMs, with the prompts for incorporating these knowledge pieces shown in \Cref{figure:prompts_all}. In \Cref{fig:kn_result}, we present the performance improvement or decline on \multihard~with \multikn~provided as external knowledge. We also plot the context window size of each model in the figure. 

\multikn~poses a significant challenge in terms of the necessary window size and the capacity to handle lengthy contexts. It is observed that models equipped with larger window sizes, i.e., GPT-4 series, are more likely to benefit from \multikn, whereas there is a notable decline in performance for MLLMs with smaller window sizes. The increase in tokens may also present a hurdle for models, as those concise questions may become overshadowed or distracted by the extensive context.

\subsection{Do these images provide information?}
\label{sec:ablation}
    \begin{table*}[!ht]
        \footnotesize
        \centering
        \setlength\tabcolsep{3.9pt}
        \renewcommand{\arraystretch}{1.0}
        \begin{tabular}{lccccccccc}
            \toprule
            \multirow{2.5}{*}{\textbf{Model}} & \multirow{2.5}{*}{\textbf{NI}} & \multicolumn{4}{c}{\textbf{SI}} & \multicolumn{4}{c}{\textbf{MI}} \\
            \cmidrule(rl){3-6}\cmidrule(rl){7-10}
             & & \textbf{w/o.} & \textbf{w. cap} & \textbf{w. ocr} & \textbf{w. img} & \textbf{w/o.} & \textbf{w. cap} & \textbf{w. ocr} & \textbf{w. img}\\
            \midrule
            MOSS          & 36.1             & 27.3             & 27.3 ($+$0.0)      & 27.6 ($+$0.3)      & -                & 17.1             & 20.7 ($+$3.6)      & 19.0 ($+$1.9)      & -                \\
            DFM-2.0       & 63.0             & 28.7             & 30.2 ($+$1.5)      & 33.4 ($+$4.7)      & -                & 11.3             & 15.6 ($+$4.3)      & 14.9 ($+$3.6)      & -                \\
            Qwen2         & 61.6             & 35.3             & 31.6 ($-$3.7)      & 35.0 ($-$0.3)      & -                & 18.0             & 16.2 ($-$1.8)      & 17.1 ($-$0.9)      & -                \\
            InternLM2.5   & 70.3             & 39.2             & 38.5 ($-$0.7)      & 38.5 ($-$0.7)      & -                & 24.5             & 28.6 ($+$4.1)      & 23.3 ($-$1.2)      & -                \\
            \hdashline
            \quad Average &                  &                  & \R{($-$0.73)}      & \G{($+$1.00)}      & -                 &                  & \G{($+$2.55)}       & \G{($+$0.85)}     & -                \\
            \midrule
            VisualGLM     & 35.1             & 20.8             & 21.4 ($+$0.6)      & 20.4 ($-$0.4)      & 25.2 ($+$4.4)      & 15.3             & 15.1 ($-$0.2)      & 14.5 ($-$0.8)      & 9.7 ($-$5.6)       \\
            VisCPM        & 36.8             & 27.1             & 27.6 ($+$0.5)      & 27.2 ($+$0.1)      & 28.4 ($+$1.3)      & 24.8             & 21.6 ($-$3.2)      & 20.9 ($-$3.9)      & 16.6 ($-$8.2)      \\
            Chinese-LLaVA & 32.3             & 26.1             & 26.3 ($+$0.2)      & 25.5 ($-$0.6)      & 22.6 ($-$3.5)      & 17.6             & 19.9 ($+$2.3)      & 19.6 ($+$2.0)      & 17.8 ($+$0.2)      \\
            Qwen-VL       & 43.2             & 30.7             & 30.3 ($-$0.4)      & 31.0 ($+$0.3)      & 32.7 ($+$2.0)      & 25.5             & 25.0 ($-$0.5)      & 26.2 ($+$0.7)      & 20.7 ($-$4.8)      \\
            Yi-VL-34B     & 63.8             & 39.9             & 38.7 ($-$1.2)      & 39.4 ($-$0.5)      & 42.0 ($+$2.1)      & 24.1             & 26.5 ($+$2.4)      & 24.2 ($+$0.1)      & 24.5 ($+$0.4)      \\
            InternVL-1.1  & 50.9             & 33.4             & 33.3 ($-$0.1)      & 33.1 ($-$0.3)      & 35.5 ($+$2.1)      & 24.8             & 21.9 ($-$2.9)      & 22.9 ($-$1.9)      & 25.1 ($+$0.3)      \\
            InternVL2-8B  & 78.7             & 48.4             & 49.0 ($+$0.6)      & 48.2 ($-$0.2)      & 51.6 ($+$3.2)      & 32.7             & 32.6 ($-$0.1)      & 28.6 ($-$4.1)      & 28.7 ($-$4.0)      \\
            MiniCPM-V-2.6 & 63.8             & 37.9             & 35.0 ($-$2.9)      & 38.3 ($+$0.4)      & 41.8 ($+$3.9)      & 21.8             & 25.1 ($+$3.3)      & 24.0 ($+$2.2)      & 24.0 ($+$2.2)      \\
            Qwen2-VL-7B   & 78.1             & 47.8             & 48.9 ($+$1.1)      & 48.5 ($+$0.7)      & 53.7 ($+$5.9)      & 29.2             & 30.4 ($+$1.2)      & 30.5 ($+$1.3)      & 34.2 ($+$5.0)      \\
            Qwen2-VL-72B  & 86.9             & 52.1             & 56.6 ($+$4.5)      & 56.6 ($+$4.5)      & 61.5 ($+$9.4)      & 28.4             & 33.0 ($+$4.6)      & 32.0 ($+$3.6)      & 41.1 ($+$12.7)      \\
            \hdashline
            \quad Average &                  &                  & \G{($+$0.29)}     & \G{($+$0.40)}       & \G{($+$3.08)}      &                  & \G{($+$0.69)}      & \R{($-$0.08)}      & \R{($-$0.18)}      \\
            \midrule
        Gemini/Vision Pro & 62.5             & 36.2             & 36.9 ($+$0.7)      & 38.4 ($+$2.2)      & 40.0 ($+$3.8)      & 18.3             & 23.2 ($+$4.9)      & 18.6 ($+$0.3)      & 24.5 ($+$6.2)      \\
            ChatGPT       & 54.0             & 6.8              & 9.9 ($+$3.1)       & 6.6 ($-$0.2)       & -                  & 5.1              & 10.7 ($+$5.6)      & 5.5 ($+$0.4)       & -                  \\
            GPT-4/V       & 74.5             & 11.3             & 9.7 ($-$1.6)       & 1.9 ($-$9.4)       & 46.9 ($+$35.6)     & 8.8              & 9.4 ($+$0.6)       & 3.1 ($-$5.7)       & 28.1 ($+$19.3)     \\
            GPT-4o-mini   & 67.7             & 17.5             & 30.0 ($+$12.5)     & 29.6 ($+$12.1)     & 43.0 ($+$25.5)     & 8.2              & 14.9 ($+$6.7)      & 10.3 ($+$2.1)      & 30.4 ($+$22.2)     \\
         Gemini 1.5 Flash & 73.8             & 26.3             & 38.7 ($+$12.4)     & 38.7 ($+$12.4)     & 50.9 ($+$24.6)     & 11.9             & 21.4 ($+$9.5)      & 15.5 ($+$3.6)     & 40.0 ($+$28.1)     \\
            GPT-4o (0806) & 80.0             & 18.5             & 33.6 ($+$15.1)     & 33.8 ($+$15.3)     & 54.1 ($+$35.6)     & 12.0             & 17.8 ($+$5.8)      & 13.2 ($+$1.2)      & 40.9 ($+$28.9)     \\
           Gemini 1.5 Pro & 80.2             & 29.4             & 43.5 ($+$14.1)     & 47.6 ($+$18.2)     & 58.5 ($+$29.1)     & 11.6             & 27.9 ($+$16.3)     & 25.4 ($+$13.8)     & 50.0 ($+$38.4)     \\
            \hdashline 
            \quad Average &                  &                  & \G{($+$5.96)}      & \G{($+$4.00)}      & \G{($+$25.13)}     &                  & \G{($+$4.72)}      & \R{($-$0.34)}      & \G{($+$19.15)}     \\
            \bottomrule
        \end{tabular}
        \caption{\textbf{Performance (\%) with different image information on \Multi.} w/o.: No image information is provided. w. cap: Image captions are provided. w. ocr: Image OCR transcripts are provided. w. img: Original images from \multi~are provided. We calculate the average improvements for each group of models and different image types.}
        \label{table:caption_result}
    \end{table*}

To assess the necessity of images in \multi~for solving problems, we conduct an ablation study where we either remove images from the SI and MI sets or substitute them with textual descriptions, such as captions and OCR-derived text. We utilize BLIP2~\cite{li2023blip} for generating image captions and EasyOCR\footnote{\url{https://pypi.org/project/easyocr/}} to extract text from images. The results are shown in \Cref{table:caption_result}.

For questions that incorporate a single image (as indicated in the SI column), the presence of images significantly aids in answering the questions, with an average performance boost of 3.08\% and 25.13\% of open-sourced MLLMs and closed-source MLLMs. Notably, both GPT-4V and GPT-4o (0806) experience a substantial increase of 35.6\% in performance, primarily due to their tendency to abstain from answering in the absence of images. We will further discuss this in \Cref{sec:reject}.

For questions that involve multiple images (as discussed in the MI column), we categorize models into three groups: 1) Closed-source models, specifically GPT-4V and Gemini Pro, which leverage all images and achieve significant improvement. 2) Open-source models capable of handling multiple high-resolution images at a time, namely Qwen2-VL and MiniCPM-V-2.6, which show promising improvements. 3) Open-source models without multi-image support, like VisualGLM, and VisCPM, which show a huge performance decline. Only one of the images given could be distracting, especially in a situation where each choice contains an image. 


In settings where images are omitted and replaced by their textual descriptions (captions or OCR text), a small improvement is observed with either captions or OCR text for most models. Captions, which generally summarize the images, introduce bilingual elements to the models, and usually miss details. OCR text, while detailed, lacks spatial information and is not universally applicable, as some images contain no text at all. Both forms of textual information lower the models' refusal rate, and LLMs benefit more from these than MLLMs. However, they potentially complicate reasoning processes. Nevertheless, a generic caption is found to more frequently outperform scattered OCR fragments.

\subsection{Will these models refuse to answer if the image is not provided?}
\label{sec:reject}
\vspace{-0.5em}
\begin{figure}[!h]
\centering
\begin{minipage}[c]{0.48\textwidth}
\centering
\includegraphics[width=1.0\linewidth,trim=0 10 0 10,clip]{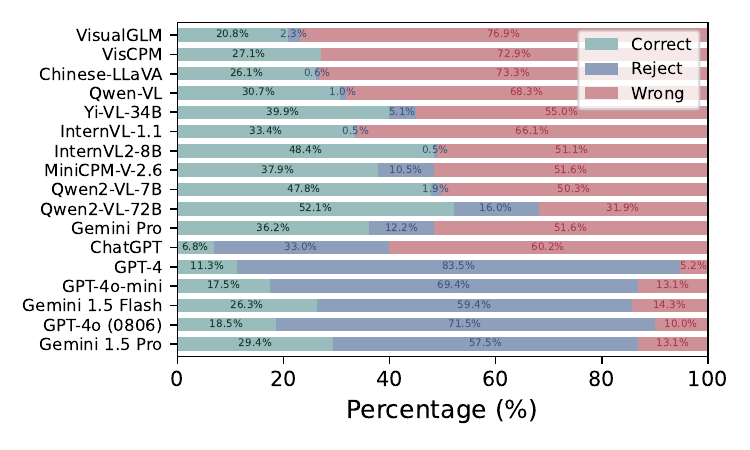}
\end{minipage}
\hspace{0.02\textwidth}
\begin{minipage}[c]{0.48\textwidth}
\centering
\includegraphics[width=1.0\linewidth,trim=0 10 0 10,clip]{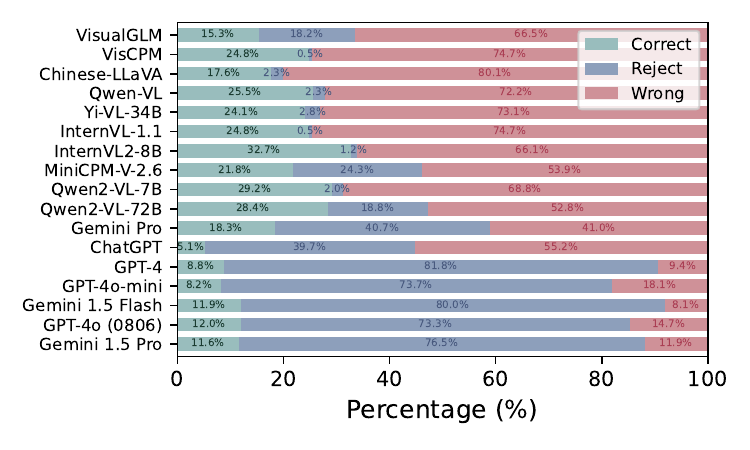}
\end{minipage}
\caption{\textbf{The portion of answer rejection when evaluated on SI set (left) and MI set (right) without image information as input.} }
\label{fig:reject}
\end{figure}

\multi~not only expects models to achieve higher scores when provided with multimodal information but also aims to investigate the performance and reliability of MLLMs under pure text input conditions. As a result, we observe that some models like the GPT series, which perform significantly better in w. img setting than pure text setting in \Cref{table:caption_result}.

We further analyzed the outputs of these models as we prompted them to reject if the information given was not sufficient. The details are presented in \Cref{fig:reject} and are within our expectations. Theoretically, the greater the performance improvement of a model with and without image information, the better it demonstrates superiority in human alignment and control over hallucinations. Strong models like GPT4 and Gemini 1.5 reject a big proportion of questions, indicating the ability to recognize missing information in the question content, which leads to a lower score. These models keep improving with user feedback and are better aligned with human preferences. Conversely, those open-sourced models are less likely to reject such questions and show a non-negligible direction for future improvement.

\subsection{Is \Multi~sensitive to different prompts?}
\label{sec:prompt_ablation}
To investigate the sensitivity of \multi~to variations in prompt settings, we conducted an ablation study using multiple prompt configurations. Specifically, we examined the following scenarios: (1) BASE: the original prompt, (2) ENG: using English as the prompt language, (3) NO BG: removing the question background in the beginning, i.e., \texttt{You need to use your knowledge of \{knowledge\} to answer this \{question\_type\} question}, (4) COT: adding zero-shot chain-of-thought\cite{cot} reasoning instructions. The results are presented in \Cref{table:ablation_prompts}.

The results indicate that \multi~is stable by switching to another language or dropping background information, yet will benefit from chain-of-thoughts reasoning.

Switching to English prompts (ENG) resulted in a minor average improvement of 1.0\%, suggesting that prompt language has limited but non-negligible influence. This might be due to a higher proportion of English fine-tuning data, which better aligns the models with English instructions. Removing the background knowledge section (NO BG) had negligible effects on the results.

Incorporating chain-of-thought reasoning (COT) consistently improves the results for most models, with an average improvement of 2.9\%. Notably, GPT-4o-mini showed the most significant benefit from COT, achieving a 12.2\% increase. The 8.9\% sharp drop in performance for Qwen2-VL-7B can be attributed to its failure to follow to the chain-of-thought prompt, where it directly provided predictions without intermediate reasoning steps in approximately 40\% of the cases. This suggests a limitation in the model's alignment with instruction-following for reasoning-intensive tasks. This trend is also observed in Qwen-VL-72B where 5\% of the questions are answered directly.


These findings demonstrate the critical role of well-crafted prompts in optimizing MLLM performance on \multi. However, using different prompt settings does not affect the overall ranking of the models in the experiments for most cases.

    \begin{table}[h]
        \begin{minipage}[c]{0.6\textwidth}
    \footnotesize
    \centering
    \renewcommand{\arraystretch}{1.0}

 \begin{tabular}{lcccc}
     \toprule
    \textbf{Model}                                         & \textbf{BASE} & \textbf{ENG}  & \textbf{NO BG} & \textbf{COT}            \\ 
    \midrule
InternVL2-8B                                           & 36.3          & 35.6 ($-$0.7) & 35.6 ($-$0.7)  & 38.3 ($+$2.0)           \\
MiniCPM-V-2.6                                          & 21.3          & 25.0 ($+$3.7) & 21.6 ($+$0.3)  & 26.3 ($+$5.0)           \\
Qwen2-VL-7B                                            & 40.0          & 40.8 ($+$0.8) & 39.0 ($-$1.0)  & 31.1 ($-$8.9)           \\
Qwen2-VL-72B                                           & 53.1          & 53.9 ($+$0.8) & 54.6 ($+$1.5)  & 51.1 ($-$2.0)           \\
GPT-4o-mini                                            & 18.5          & 20.2 ($+$1.7) & 21.2 ($+$2.7)  & 30.7 ($+$12.2)          \\
GPT-4o (1120)                                          & 36.3          & 37.6 ($+$1.3) & 38.8 ($+$2.5)  & 42.0 ($+$5.7)           \\
Gemini 1.5 Flash                                       & 33.4          & 34.0 ($+$0.6) & 33.1 ($-$0.3)  & 36.4 ($+$3.0)           \\
Gemini 1.5 Pro                                         & 46.9          & 46.1 ($-$0.8) & 45.0 ($-$1.9)  & 52.6 ($+$5.7)           \\
\hdashline
\quad Average &               & \G{($+$1.0)}      & \G{($+$0.4)}       & \G{($+$2.9)} \\ 
\bottomrule
\end{tabular}

    \caption{\textbf{The comparison of different prompt settings (\%).} BASE: the original prompt. ENG: English translation of the original prompt. NO BG: remove background information. COT: zero-shot chain-of-thought.} 
    \label{table:ablation_prompts}
        \end{minipage}
        \hspace{0.04\textwidth}
    \begin{minipage}[c]{0.34\textwidth}
        \centering
        \includegraphics[width=\textwidth,trim=0 0 0 0,clip]{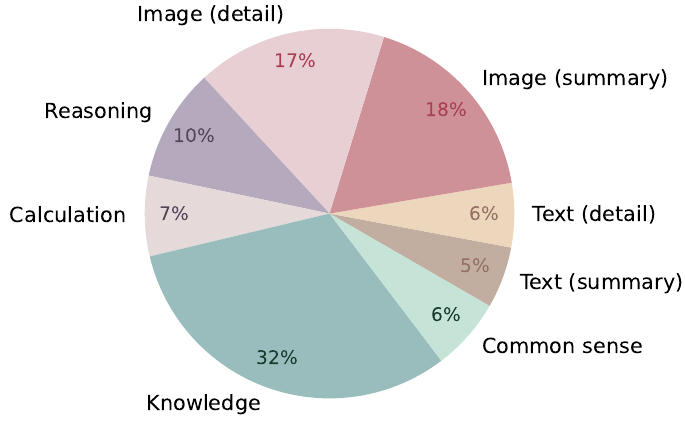}
        \figcaption{\textbf{The required ability distribution of \Multihard.}}
        \label{figure:abilities}
        \end{minipage}
    \end{table}

    \begin{figure}[!h]
        \centering
        \includegraphics[width=1.0\textwidth,trim=0 10 0 10,clip]{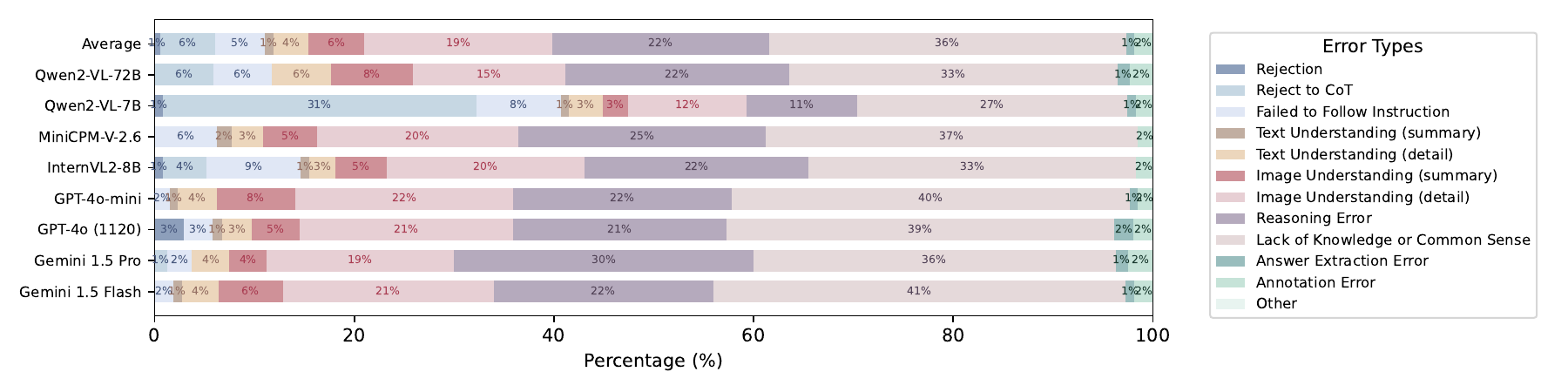}
        \caption{\textbf{The error distributions of several models' responses.}}
        \label{figure:error}
    \end{figure}
    
\subsection{What is the main cause of the error?}
\label{sec:error}

We asked human experts to identify key abilities required to solve each question, such as comprehension of text and images, mathematical calculation, logical reasoning, knowledge recall, and others. The distribution of these abilities is shown in \Cref{figure:abilities}, allowing us to compare the alignment between the skills demanded by the questions and the following observed error patterns.

To explore the primary causes of errors, we analyzed 165 questions (one-third of \multihard), using predictions generated by the 8 models evaluated in \Cref{sec:prompt_ablation}. To get more detailed insights, we focused on outputs generated with the CoT prompt, which helps reveal patterns and challenges across different error types. The detailed distribution of all errors is shown in \Cref{figure:error}, and we give five example cases with error analysis for several selected models as shown in \ref{sec:app-case}. 

We summarize these errors into six primary categories, with the following observations:

\textbf{Alignment Errors (11.1\%)}: A significant proportion of errors caused by models failing to follow CoT (Qwen2-VL-7B in \Cref{figure:case_aat} and \Cref{figure:case_phy}) or formatting instructions (InternVL2-8B in \Cref{figure:case_phy}), highlighting difficulties in instruction compliance. Some models also rejected questions outright, potentially due to divergences in human alignment objectives. Notably, Qwen2-VL-7B and 72B exhibited higher rejection rates when prompted with CoT but proceeded to provide direct answers, as discussed in \Cref{sec:prompt_ablation}.

\textbf{Text Understanding Errors (4.4\%)}: Failures in text comprehension account for 4\% to 6\% of errors, which is lower than the importance of text understanding identified by human experts. Divided into summary-level and detail-level errors, these failures typically involved an inability to extract critical context words, even though the models performed well on general understanding. Or, conversely, the model pays too much attention to detail and ignores the overall theme (GPT-4o-mini in \Cref{figure:case_hist}).

\textbf{Image Understanding Errors (24.4\%)}: Much different from text understanding, errors in image summarization and detailed analysis account for 15\% to 30\% of total errors. Since 107 (65\%) out of the 165 questions analyzed involve images, it demonstrates the challenge of handling image inputs. Similar to text errors, detail-level image comprehension caused more errors,
as it is the bottleneck for all models to handle the AAT questions, just like the one in \Cref{figure:case_aat}. There are cases in which models often struggle to locate and focus on specific image elements as mentioned in the text (GPT-4o in \Cref{figure:case_phy}), sometimes resulting in visual hallucinations. Enhanced strategies for analyzing and synthesizing image details are crucial to address this issue.

\textbf{Reasoning Errors (21.7\%)}: Logical reasoning errors emerged primarily in tasks requiring multi-step deductions (MiniCPM-V-2.6 in \Cref{figure:case_bio}) or calculations after extracting relevant information from text and image contexts. There is also the possibility that the model just repeats its ineffective reasoning steps for all options (InternVL2-8B in \Cref{figure:case_hist}). These errors underscore the difficulty of complex reasoning, which remains a significant bottleneck for both LLMs and MLLMs.

\textbf{Lack of Knowledge or Common Sense (35.8\%)}: Errors not attributed to the categories above often resulted from insufficient domain-specific knowledge. Models frequently failed to retrieve or apply relevant knowledge to solve or even comprehend the questions (GPT-4o-mini in \Cref{figure:case_aat} and \Cref{figure:case_drive}). As discussed in \Cref{sec:kn}, adding domain-specific knowledge has shown improvement for certain models, provided they can utilize the added information without being misled. Yet there is much to learn for models in training stages, and a challenge to effectively adopt retrieval augmented generation (RAG).

\textbf{System Bias (2.6\%)}: A small proportion of errors resulted from annotation inaccuracies or answer extraction issues, categorized as system bias. These errors have been addressed and corrected in the latest dataset release.

In summary, the observed error types align closely with the skill requirements identified in the dataset, reflecting the persistent challenges in reasoning and multimodal comprehension for MLLMs, even with CoT prompting. Addressing these obstacles will require significant advancements in model alignment, instruction following, and cross-modal reasoning capabilities.

\subsection{Takeaways}
    \label{sec:takehome}

    \begin{itemize}
    \item Qwen2-VL-72B demonstrates the highest performance with a 76.9\% score, indicating a significant challenge of \multi, while Gemini 1.5 Pro leads among closed-source models with 71.8\%. There is still a large gap between most capable MLLMs and human experts in general, while in some subjects like the Driving Test, Qwen2-VL-72B has surpassed the human baseline, but the AAT test still remains as the greatest challenge for all models. (\Cref{table:overall_result} in \Cref{sec:main})

    \item MLLMs show a performance drop in questions requiring more images, with only Gemini 1.5 Pro reaching 50.0\% of the score rate in multi-image scenarios. (\Cref{table:image_result} in \Cref{sec:main})
    
    \item Models generally perform better on questions requiring shorter answers, i.e., SA $>$ MA $>$ FB $>$ OP. The results of MA Acc. emphasize the importance of balancing recall and precision, where less capable MLLMs failed to provide all correct choices. (\Cref{table:type_result} in \Cref{sec:main})
    
    \item In the \multihard~evaluation, the gap between models and human experts becomes greater by 20.0\%, illustrating the difficulty of \multihard~and the need for advanced image understanding and reasoning across modalities. (\Cref{table:extreme_result} in \Cref{sec:hard})

    \item The inclusion of \multikn~is likely to help improve performance on models with longer window sizes. However, it might also have adverse effects due to multiple reasons. (\Cref{fig:kn_result} in \Cref{sec:kn})
    
    \item Including image information in questions significantly enhances model performance, especially when original images are included, though textual replacements like captions or OCR text provide some utility but lack comprehensive spatial and detail fidelity. (\Cref{table:caption_result} in \Cref{sec:ablation})

    \item Strong models like GPT-4 series and Gemini 1.5 effectively refuse to answer when critical image information is missing, showcasing better alignment with human preferences, while open-source models are likely to answer based on available text information. (\Cref{fig:reject} in \Cref{sec:reject})

    \item Prompt variations affect performance, with chain-of-thought reasoning enhancing accuracy while removing system-level instructions leads to notable declines. The relative performance across models remains stable and robust to different prompt settings. (\Cref{table:ablation_prompts} in \Cref{sec:prompt_ablation})

    \item Errors in model performance are primarily driven by failures in knowledge recall, logical reasoning, and image understanding. Additionally, alignment and instruction-following issues underscore challenges in maintaining adherence to complex prompts, particularly in chain-of-thought tasks. The observed error types match closely with the skill requirements identified by human annotators. (\Cref{figure:abilities} and \Cref{figure:error} in \Cref{sec:error})

    \end{itemize}
    
\section{Conclusion}
    In this paper, we introduce \multi, a comprehensive and challenging benchmark designed to rigorously evaluate the performance of MLLMs in detailed cross-modality understanding and scientific reasoning. Our experiments with state-of-the-art models like Qwen2-VL, InternVL2, Gemini 1.5, and GPT-4 series demonstrate that while these models exhibit promising capabilities, there remains a significant gap compared to human performance, particularly in tasks involving cross-modal alignment, logical reasoning, and complex comprehension. This underscores the need for continuous research and development in this domain.

    The creation of the \multihard~and \multikn~subsets further contributes to the field by providing insights into the strengths and limitations of current MLLMs. These subsets challenge the models' learning and reasoning abilities and encourage the development of more sophisticated and robust multimodal understanding systems.

    \multi~benchmark opens new avenues for research, particularly in enhancing the MLLMs' ability to integrate and reason over diverse data types, including images, text, and structured data. Future work may focus on expanding the benchmark to include more diverse modalities and question types, further pushing the boundaries of what MLLMs can achieve. By making \multi~publicly available, we hope to foster a collaborative environment where researchers can continuously test and improve the capabilities of MLLMs, driving the field toward the development of truly intelligent and versatile AI systems.

\Acknowledgements{This work was funded by National Natural Science Foundation of China Projects (Grant Nos. 92370206, U23B2057, and 62120106006), National Science and Technology Major Project (Grant No. 2023ZD0120703), and Shanghai Municipal Science and Technology Projects (Grant Nos. 2021SHZDZX0102 and 25X010202846). The authors would like to thank Kunyao LAN and Lei PAN for their contribution during the revision process.}


\newpage
\bibliographystyle{scis.bst}
\bibliography{custom}

\begin{appendix}

\section{Limitations and Future Work}
    \begin{itemize}
        \item \textbf{Potential biases:} While we have taken several steps to minimize biases in the dataset and ensure fair evaluation, we acknowledge that some biases remain. Despite our efforts, challenges in the construction and evaluation process, such as subject, difficulty, format, and system biases, still persist to some extent. The distribution of questions across different subjects and difficulty levels may still exhibit some inherent imbalances, which could impact the evaluation of MLLMs in specific domains or tasks. Although we have adopted algorithms for subject selection and multi-round annotation for difficulty balancing, these processes are not foolproof. Additionally, variations in the annotation process, such as subjective difficulty ratings and semantic rephrasing of questions, may introduce slight inconsistencies across different subsets of the data. We are committed to refining our methodology in future versions to further mitigate these biases and enhance the fairness of the benchmark.
        \item \textbf{Multilingual capabilities:} \multi~predominantly features simplified Chinese and mainly focuses on subjects taught in Chinese schools, with limited English multimodal content that's relatively straightforward for MLLMs. We plan to include translations in future versions. Nonetheless, the presence of Chinese characters in figures poses a significant challenge for MLLMs trained on different linguistic datasets.
        \item \textbf{Use of explanations:} While we have annotated explanations in detail, the utilization in subsequent studies remains limited. These explanations could potentially serve as valuable training data for model fine-tuning and few-shot learning using methods like CoT (Chain-of-Thoughts) or RAG (Retrieval Augmented Generation) and may aid in evaluating reasoning skills.
        \item \textbf{Metrics for evaluating blank-filling, open-ended writing, and others:} Our evaluation primarily uses exact match, which might be overly stringent for assessing MLLMs' true capabilities. Assessing open-ended writing tasks that require complex knowledge and reasoning is still a challenge. We also have nearly 100 questions that do not belong to traditional categories, such as questions requiring geographic drawing, and the evaluation of them will be even more challenging. Now that only a few studies~\cite{wang2023scibench} involve human evaluation, developing automatic and reliable methods remains an open research area.
        \item \textbf{Adaptation to various MLLMs:} Although we have tested several MLLMs, numerous others exist and new ones are continuously emerging. We encourage the community to evaluate their MLLMs using our benchmark to gauge their cognitive reasoning abilities. We will test more models as soon as the multilingual version is released.

    \end{itemize}

    \section{Several Useful Links}
    \label{sec:app-links}
    \noindent Official Website:  \url{https://opendfm.github.io/MULTI-Benchmark}
    
    \noindent Evaluation Code of \multi: \url{https://github.com/OpenDFM/MULTI-Benchmark}
    
    \noindent Prompts used in \multi: \url{https://github.com/OpenDFM/MULTI-Benchmark/blob/main/eval/prompts.py}
    
    \noindent Download \multi: \url{https://huggingface.co/datasets/OpenDFM/MULTI-Benchmark}
    
    \noindent Realtime Leaderboard: \url{https://opendfm.github.io/MULTI-Benchmark/\#leaderboard}

    \section{Models and Experiment Cost}
    \label{sec:model}
    The model specifications are listed in \Cref{table:models}. 

        \begin{table}[!h]
            \footnotesize
            \centering
            \setlength\tabcolsep{7.5pt}
            \renewcommand{\arraystretch}{1.0}
            \begin{tabular}{rlcccccc}
                \toprule
                \textbf{Creator} & \textbf{Model}                                & \textbf{Date} & \textbf{\# Paras} & \textbf{Form} & \textbf{Modality} & \textbf{Version} \\
                \midrule
                FDU              & MOSS~\cite{sun2023moss}                       & 2023-04-19    & 16 B               & Weight        & T                 & \texttt{moss-moon-003-sft}          \\
                AISpeech         & DFM-2.0~\cite{chen2022dialogzoo}              & 2024-01-16    & 70 B               & Weight        & T                 & \texttt{dfm-2.0-70b-preview}        \\
                Alibaba          & Qwen2-7B~\cite{qwen2}                         & 2024-06-04    & 7 B                & Weight        & T                 & \texttt{Qwen2-7B-Instruct}          \\
                PJLab            & InternLM2.5-7B~\cite{cai2024internlm2}        & 2024-06-27    & 7 B                & Weight        & T                 & \texttt{internlm2\_5-7b-chat}       \\
                \hdashline
                THU              & VisualGLM~\cite{du2022glm}                    & 2023-05-17    & 6 B                & Weight        & SI                & \texttt{visualglm-6b}               \\
                OpenBMB          & VisCPM~\cite{viscpm}                          & 2023-06-28    & 10 B               & Weight        & \underline{SI}    & \texttt{VisCPM-Chat}                \\
                LinkSoul-AI      & Chinese-LLaVA~\cite{cllava}                   & 2023-06-30    & 7 B                & Weight        & One               & \texttt{Chinese-LLaVA-Cllama2}      \\
                Alibaba          & Qwen-VL~\cite{bai2023qwen}                    & 2023-09-25    & 7 B                & Weight        & MI                & \texttt{Qwen-VL-Chat}               \\
                01-ai            & Yi-VL-34B~\cite{yi}                           & 2023-11-22    & 34 B               & Weight        & \underline{One}   & \texttt{Yi-VL-34B}                  \\
                OpenGVLab        & InternVL-1.1~\cite{chen2023internvl}          & 2024-01-24    & 19 B               & Weight        & \underline{One}   & \texttt{InternVL-Chat-V1-1}         \\
                OpenGVLab        & InternVL2-8B~\cite{chen2024far}               & 2024-07-03    & 8 B                & Weight        & MI                & \texttt{InternVL2-8B}               \\
                OpenBMB          & MiniCPM-V-2.6~\cite{yao2024minicpm}           & 2024-08-05    & 8 B                & Weight        & MI                & \texttt{MiniCPM-V-2\_6}             \\
                Alibaba          & Qwen2-VL-7B~\cite{wang2024qwen2}              & 2024-08-28    & 7 B                & Weight        & MI                & \texttt{Qwen2-VL-7B-Instruct}       \\
                Alibaba          & Qwen2-VL-72B~\cite{wang2024qwen2}             & 2024-08-28    & 72 B               & Weight        & MI                & \texttt{Qwen2-VL-72B-Instruct}      \\
                \midrule
                OpenAI           & ChatGPT~\cite{chatgpt}                        & 2023-11-06    & -                 & API           & T                 & \texttt{gpt-3.5-turbo-1106}         \\
                OpenAI           & GPT-4~\cite{gpt4}                             & 2023-11-06    & -                 & API           & T                 & \texttt{gpt-4-1106-preview}         \\
                Google           & Gemini Pro~\cite{geminiteam2023gemini}        & 2023-12-07    & -                 & API           & T                 & \texttt{gemini-1.0-pro-001}         \\
                \hdashline
                OpenAI           & GPT-4V~\cite{gpt4v}                           & 2023-11-06    & -                 & API           & MI                & \texttt{gpt-4-1106-vision-preview}  \\
                Google           & Gemini Vision Pro~\cite{geminiteam2023gemini} & 2023-12-07    & -                 & API           & \underline{MI}    & \texttt{gemini-1.0-pro-vision}      \\
                OpenAI           & GPT-4o-mini~\cite{gpt4omini}                  & 2024-07-18    & -                 & API           & MI                & \texttt{gpt-4o-mini-2024-07-18}     \\
                OpenAI           & GPT-4o (0806)~\cite{gpt4o}                    & 2024-08-06    & -                 & API           & MI                & \texttt{gpt-4o-2024-08-06}          \\
                Google           & Gemini 1.5 Flash~\cite{geminiteam2023gemini}  & 2024-08-27    & -                 & API           & MI                & \texttt{gemini-1.5-flash-exp-0827}  \\
                Anthropic        & Claude 3.5 Sonnet~\cite{claude3}              & 2024-10-22    & -                 & API           & MI                & \texttt{claude-3-5-sonnet-20241022} \\
                OpenAI           & GPT-4o (1120)~\cite{gpt4o}                    & 2024-11-20    & -                 & API           & MI                & \texttt{gpt-4o-2024-11-20}          \\
                Google           & Gemini 1.5 Pro~\cite{geminiteam2023gemini}    & 2024-11-21    & -                 & API           & MI                & \texttt{gemini-exp-1121}            \\
                \bottomrule
            \end{tabular}
            \caption{\textbf{The list of models evaluated on \Multi.} We report Modality as how many images can the model take in one turn. Note that those MLLMs commonly support multiple-image input with chatting in several turns. T: pure text LLM, One: only one image in the beginning, SI: single image in each turn, MI: multiple images in one turn. The \underline{underline} means the model must have an image as input. }
            \label{table:models}
        \end{table}

A standard evaluation using GPT-4o with image inputs (without CoT reasoning) requires approximately 4.5 million prompt tokens and 150k completion tokens, amounting to a total cost of around \$25. Running on a single A100 (80 G) GPU, models with 7–8 B parameters complete the benchmark in 1–2 h, while a 70 B+ model takes approximately 8 h on 2-4 A100 GPUs. CoT experiments demand significantly more resources due to the larger token output. To facilitate more practical and faster usage, we recommend researchers prioritize testing on \multihard, which requires only around \$1 and a few minutes to complete.

    \section{Prompts}
    \label{sec:prompts}

\begin{figure*}[!h]
        \centering
        \includegraphics[width=1.0\textwidth,trim=5 150 5 5,clip]{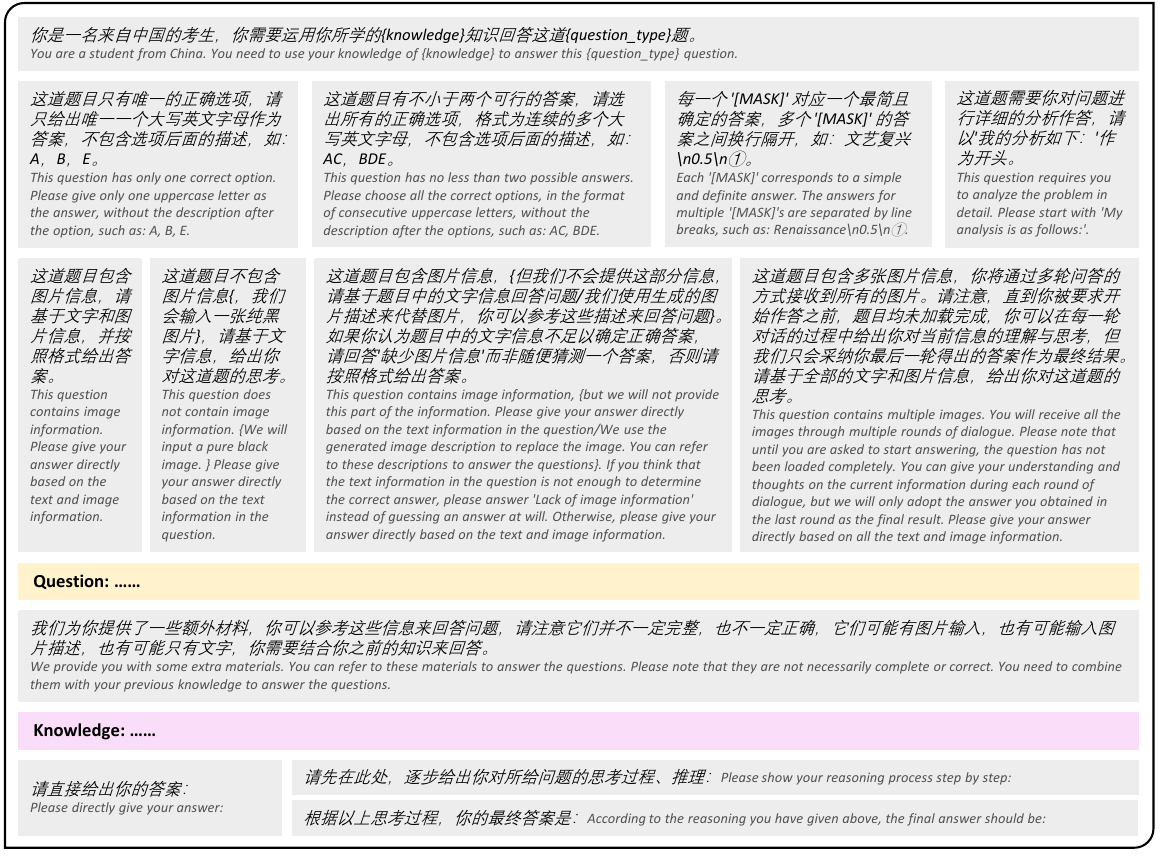}
        \caption{\textbf{The prompts for evaluation on \Multi.}}
        \label{figure:prompts_all}
    \end{figure*}

    The complete collection of prompts designed for evaluation on \multi~is shown in \Cref{figure:prompts_all}. One of the prompt pieces in each row is selected according to the evaluation setting and data format. Please note that some prompts will not take effect under certain cases, for instance, the prompt related to knowledge will be omitted if the knowledge is not given.

    \section{Data Selection Algorithm}
    \label{sec:algo}

    We mostly pick questions based on its content length $L_q$, calculated with function

    $$
    \begin{aligned}
        L_q=\left(a \times \begin{bmatrix}
                                \mathcal{H}(L_{q,\# \text{characters in question}}) \\
                                \mathcal{H}(L_{q,\# \text{characters in answer}})   \\
                                \mathcal{H}(L_{q,\# \text{characters in analysis}})
        \end{bmatrix}\right.+ \left.b \times \begin{bmatrix}
                            \mathcal{H}(L_{q,\# \text{images in question}}) \\
                            \mathcal{H}(L_{q,\# \text{images in answer}})   \\
                            \mathcal{H}(L_{q,\# \text{images in analysis}})
        \end{bmatrix} \right)^\top\begin{bmatrix}
                                      1.0 \\
                                      0.1 \\
                                      0.5
        \end{bmatrix}
    \end{aligned}
    $$
    where $a=1,b=1$ are customized weights.

    In the formula above, we use a harmonic mean function $\mathcal{H}$ to normalize content length $L_{q,i}$ of each target value $i$ within each knowledge piece $k$.\footnote{Note that for those questions without knowledge information, we simply use an empty string as a keyword.}

    $$
    \mathcal{H}(L_{q,i})=\frac{1}{\frac{1}{L_{q,i}}+\frac{1}{\overline{L_{q,i}}}}=\frac{2L_{q,i}\overline{L_{q,i}}}{L_{q,i}^2+\overline{L_{q,i}}^2}
    $$
    where $\overline{L_{q,i}}$ is the arithmetic average of $L_{q,i}$ for all questions with $k$.

    Then we pick $N_k$ questions within each knowledge piece $k$.
    
    $$
    N_k= \lceil \alpha \times \lg( \# \text{questions of } k) \rceil
    $$
    where $\alpha=3$ is a customized parameter.
    
    Now we sort $L_{q,k}={L_q: q\in k}$ in descendent order.
    
    Then we assign a pick-up probability to select these questions
    $$
    \text{Pr}[\text{pick up }q]=
    \begin{cases}
        1&\text{ , for }q \text{ s.t. } L_{q,k}[0]  \\
        p&\text{ , if }q=1\text{ , for }q \text{ of } L_{q,k}[1:m] \text{ or } L_{q,k}[-m:]\\
        p\frac{N_k-2m}{\# \text{questions of } k}&\text{ , otherwise}
    \end{cases}
    $$

\section{Data Process and Annotation}
\label{sec:app-data}

    Initially, we extract a total of 2.7 million questions from the Internet. Through an algorithmic selection in the preprocessing stage, we narrow this down to 18,000 questions with wide coverage. During the construction, we conduct two rounds of data annotation and three rounds of automatic checking to ensure the granularity and credibility of every question in our set. In the first round of annotation, we filter out and modify questions based on predefined criteria. The second round of data annotation focuses more on semantic analysis and data enhancement. This post-processing stage significantly increases the number of MA questions by 3.22 times, and the total point proportion of non-SA questions rose from 26.0\% to 40.1\%. We also remove over 800 similar questions. 

    \subsection{Data Pre-process}
    The raw data range from HTML, photocopy, hand script, and plain text, and we conduct pre-processing to reduce the load of further annotation. We remove most HTML tags indicating irrelevant content of the question, such as alignment, color, etc. We reserve tags for underlines (\texttt{<u> </u>}), and we transfer several tagged styles, including bold, italic, and tabular data, into markdown format. For some complex tables that cannot be well converted, we save them as a screenshot picture after rendering with HTML.
    
    For photocopy and hand script, we adopt OCR tools to convert text content, crop images, and figures, and integrate them into markdown. We further transcribe most of the math functions and chemistry structures into \LaTeX~format, with a small portion remaining as images. 

    \subsection{Data Annotation}

    \begin{figure*}[!h]
        \centering
        \includegraphics[width=0.9\textwidth,trim=150 30 150 50,clip]{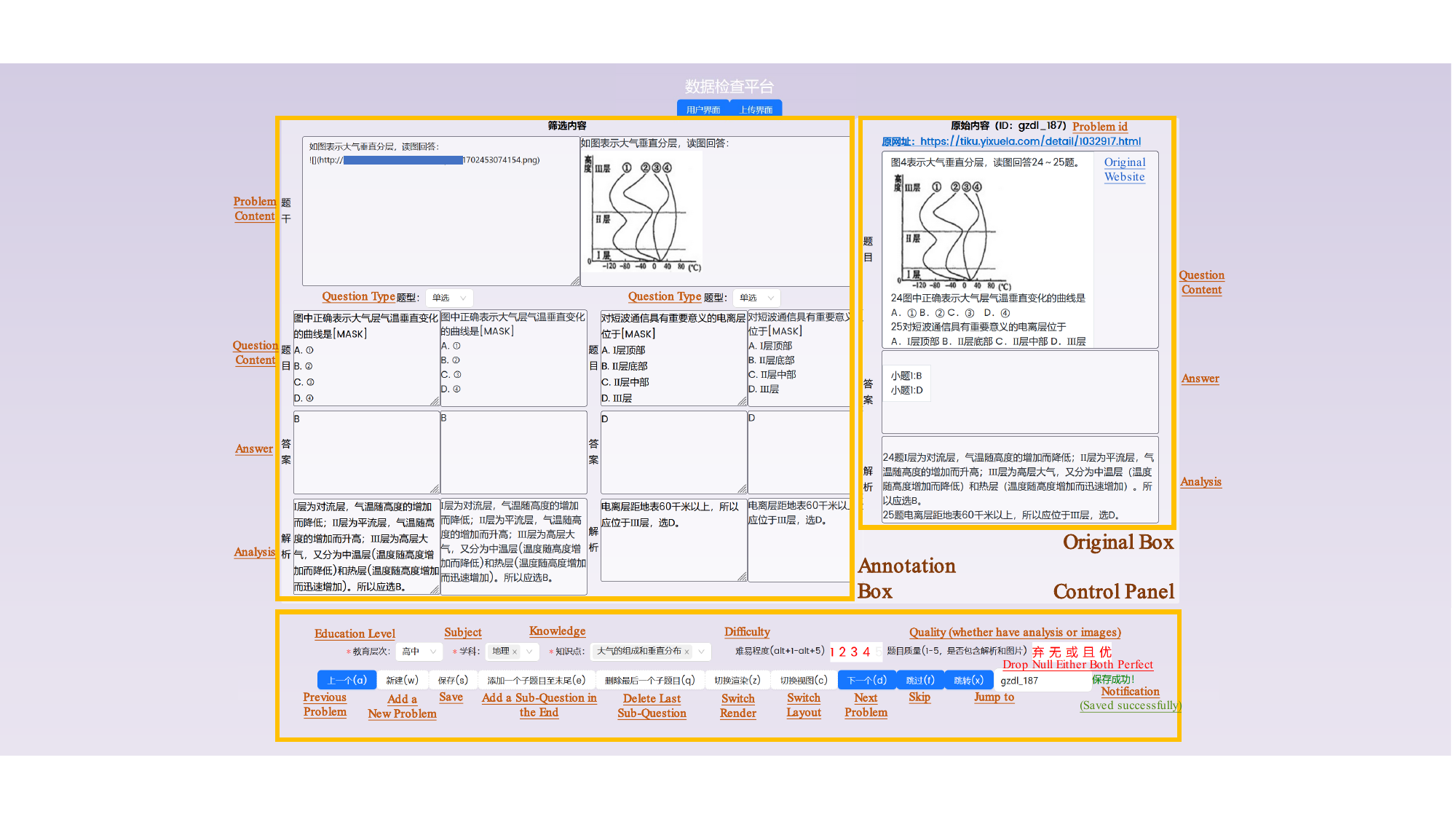}
        \caption{\textbf{A screenshot of the main page of the data annotation platform.}}
        \label{figure:platform}
    \end{figure*}
    
\begin{figure*}[!th]
        \centering
        \includegraphics[width=0.9\textwidth,trim=5 170 5 5,clip]{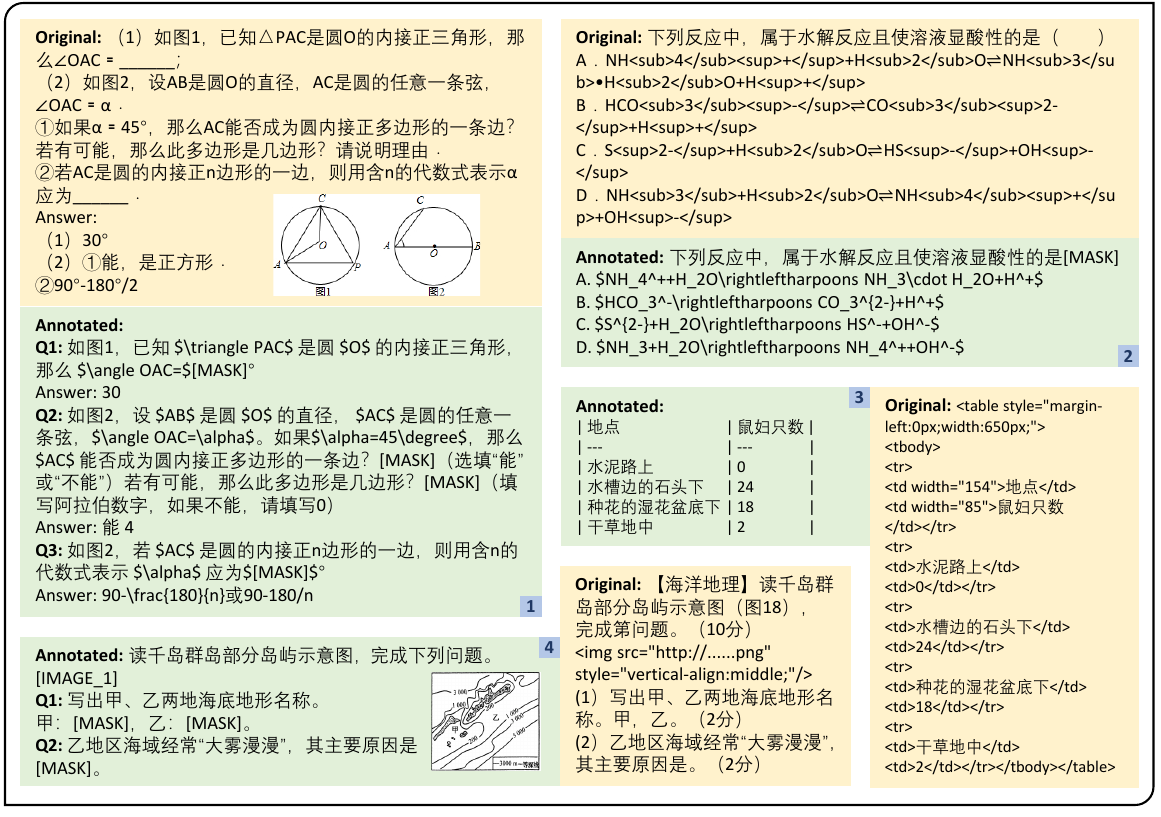}
        \caption{\textbf{Several data annotation examples when constructing \Multi.} }
        \label{figure:data_anno}
    \end{figure*}

    We develop an online platform for the data annotation stage. The platform consists of text boxes for editing contents and regions for rendering the text to see the final appearance of the data as shown in \Cref{figure:platform}. We employ skilled human annotators and involve them as authors, primarily undergraduate students from top universities in China familiar with exam quizzes and markdown rules, to undertake this comprehensive task covering various aspects from formatting to semantic analysis:
    \begin{itemize}
        \item \textbf{Format Level}. Tasks at this level involve the removal of residual HTML tags and the conversion of content into markdown format (refer to examples (1) and (3) in \Cref{figure:data_anno}). This includes transforming complex mathematical and chemical equations, usually in HTML, into \LaTeX. For this purpose, Mathpix \footnote{\url{https://mathpix.com/snipping-tool}} is utilized for efficiency. The annotators also correct any character-level errors in text and formulas, often resulting from OCR inaccuracies. 

        \item \textbf{Content Level}. Annotators split the raw content into distinct sub-questions, segregating parts like the question, answer, and analysis (if presented in raw data). We divide the question content into general and specific parts. The general part includes the problem introduction, background information, or instructions applicable across all sub-questions, while the specific part contains details unique to each sub-question. Annotators also standardize typesetting and image placement, ensuring a consistent format across questions of the same type (e.g., for multiple-choice questions with a single image, the format follows \texttt{problem content(general) + question content(specific) + [MASK] + [IMAGE\_1] + choices}).

        \item \textbf{Label Level}. Annotators evaluate each question's difficulty and quality. A question is considered of higher quality if it includes comprehensive content, multiple images, or detailed explanations. Difficulty assessment is subjective. These evaluations aid in curating our \multihard~dataset. Annotators also verify information like question type, educational level, and related knowledge pieces.

        \item \textbf{Semantic Level}. At this stage, annotators are advised to identify and correct both superficial errors (e.g., empty/duplicate choices, incomplete mathematical functions such as between \texttt{\$32\$}, \texttt{\$3\^{}2\$}, \texttt{\$\textbackslash sqrt[3]\{2\}\$}, \texttt{\$3\textbackslash sqrt\{2\}\$}, \texttt{\$\textbackslash frac\{3\}\{2\}\$}) and more profound errors relating to factual accuracy and logical reasoning, such as content that is lacking or leads to inconclusive results. Those questions with profound errors will be dropped.
    \end{itemize}

    In \Cref{figure:data_anno}, we show several examples of complex formation and modification during the data annotation stage. The markdown, \LaTeX, and HTML format code remains for better formatting clarity.

    \subsection{Data Post-process}
\begin{figure}[h]
        \centering
        \includegraphics[width=0.9\textwidth,trim=5 235 5 5,clip]{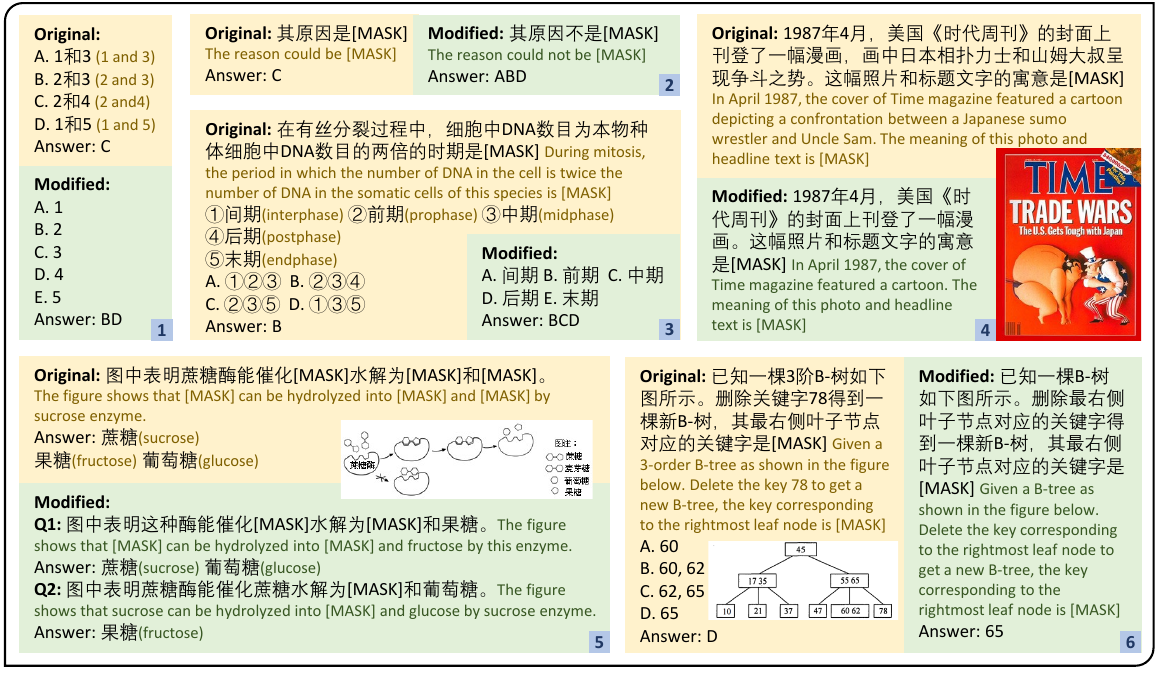}
        \caption{\textbf{Several data augmentation examples when constructing \Multi.}}
        \label{figure:data_aug}
    \end{figure}

    To collect more challenging data for our benchmark, we adopt several data post-processing strategies:
    \begin{itemize}
        \item \textbf{Formation}. During the data preprocessing stage and annotation stage, we format the questions in a render-friendly manner, and meanwhile reduce the similarity to contents that the MLLMs are trained on. During this stage, we assess if there are any omissions or missing elements.

        \item \textbf{Disambiguration}. For blank-filling questions containing multiple \texttt{[MASK]}s, we manually modify those with parallel relations into two sub-questions (refer to example (5) in \Cref{figure:data_aug}) to determine a unique fixed answer for each question.

        \item \textbf{Distillation}. This is completed during our annotation process. We reduce assistance information so that the answer must depend on a more detailed analysis (refer to example (4) in \Cref{figure:data_aug}). In this way, we greatly enhance question difficulty.

        \item \textbf{Transformation}. We randomly modify the questions, such as from single-choice to blank-filling (refer to example (2) in \Cref{figure:data_aug}), or convert certain kinds of single-choice questions into multiple-choice ones (refer to example (1) and (5) in \Cref{figure:data_aug}). Lots of single-choice questions have a list of options, and the choices are presented as a combination of those options where only one is correct. We transform those questions into multiple-choice questions where the options become new choices and the correct answer corresponds to the combinations. In this way, we successfully increase the scale of multiple-choice questions, improving the diversity of the questions.
    \end{itemize}

  In \Cref{figure:data_aug}, we show several examples of complex formation and modification during the data postprocess stage. English translations of Chinese text are shown for better readability.

\begin{figure}[h]
        \centering
        \includegraphics[width=0.9\textwidth,trim=5 145 5 5,clip]{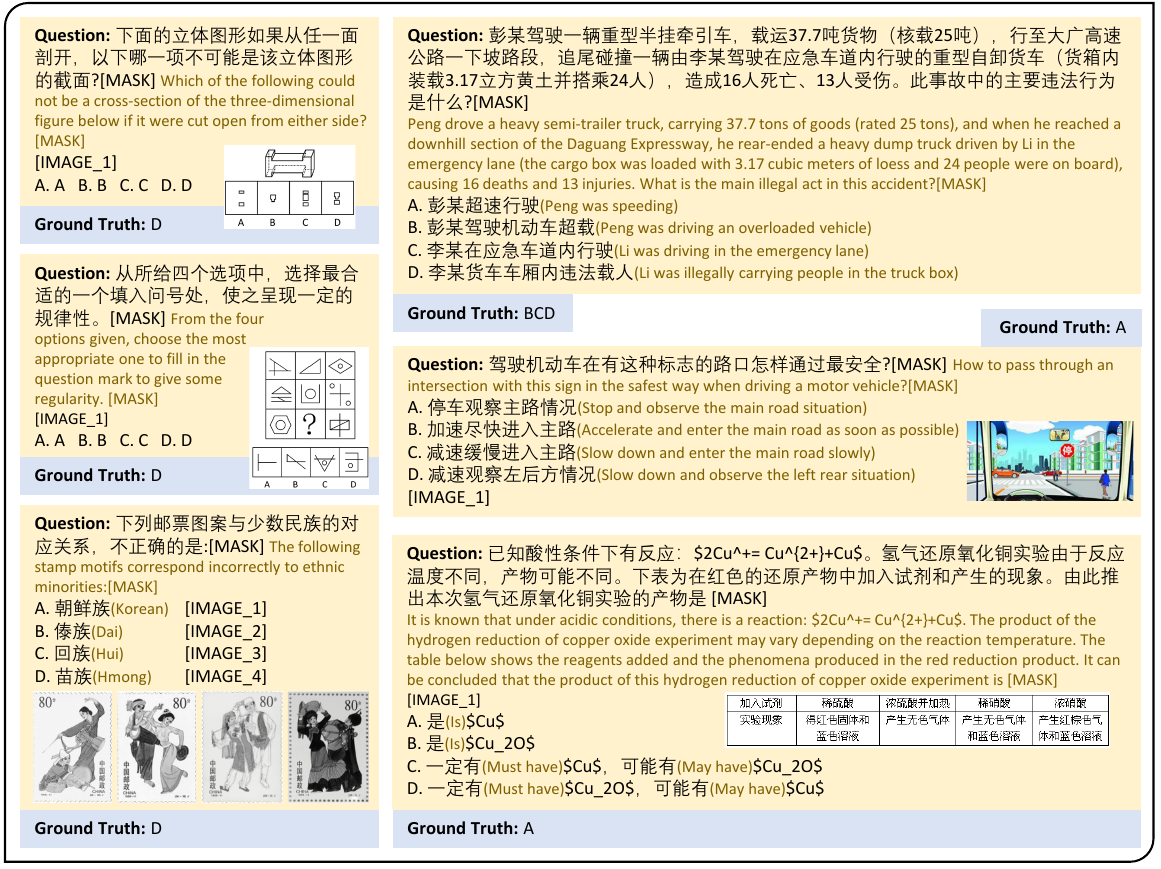}
        \caption{\textbf{More examples of \Multi.}}
        \label{figure:examples}
    \end{figure}

     In \Cref{figure:examples}, we show more examples for annotated questions. English translations of Chinese text are shown for better readability.

\clearpage
    \section{Case study}
\label{sec:app-case}

    In this section, we provide 5 case studies alone with annotations by human experts and error analysis for several selected models.

    \begin{figure}[!h]
        \centering
        \includegraphics[width=0.9\textwidth,trim=5 215 5 5,clip]{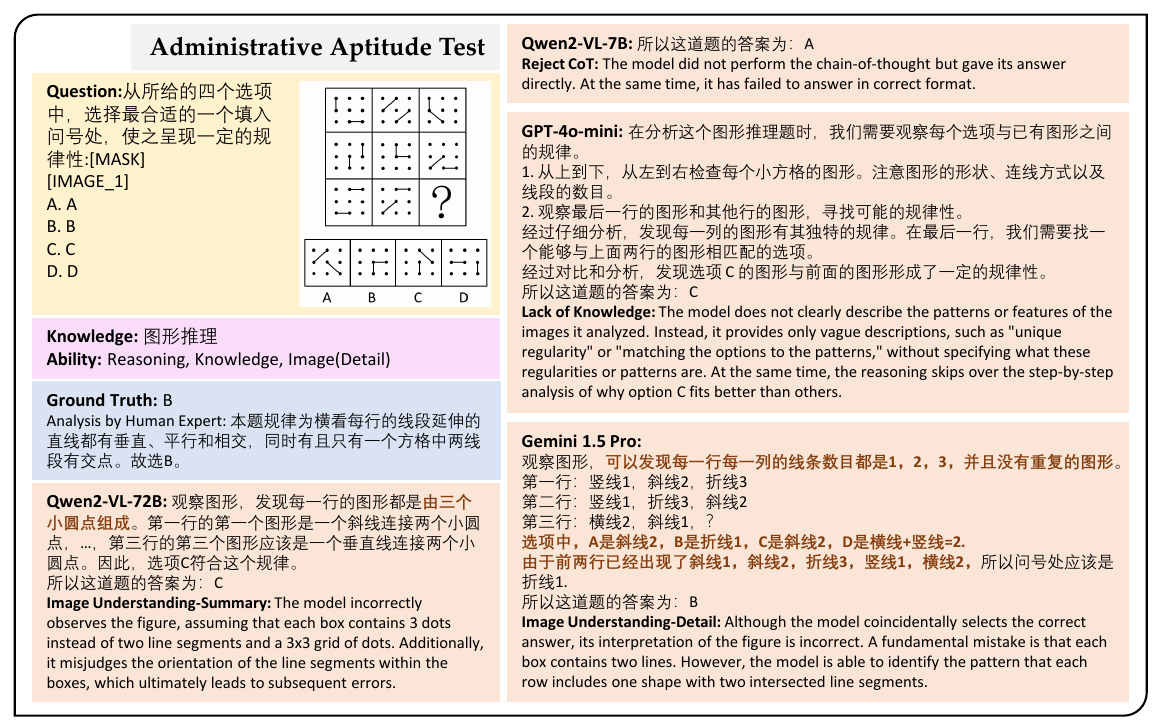}
        \caption{\textbf{Error analysis of an SI-SA question.}}
        \label{figure:case_aat}
    \end{figure}
    
    \begin{figure}[!h]
        \centering
        \includegraphics[width=0.9\textwidth,trim=5 270 5 5,clip]{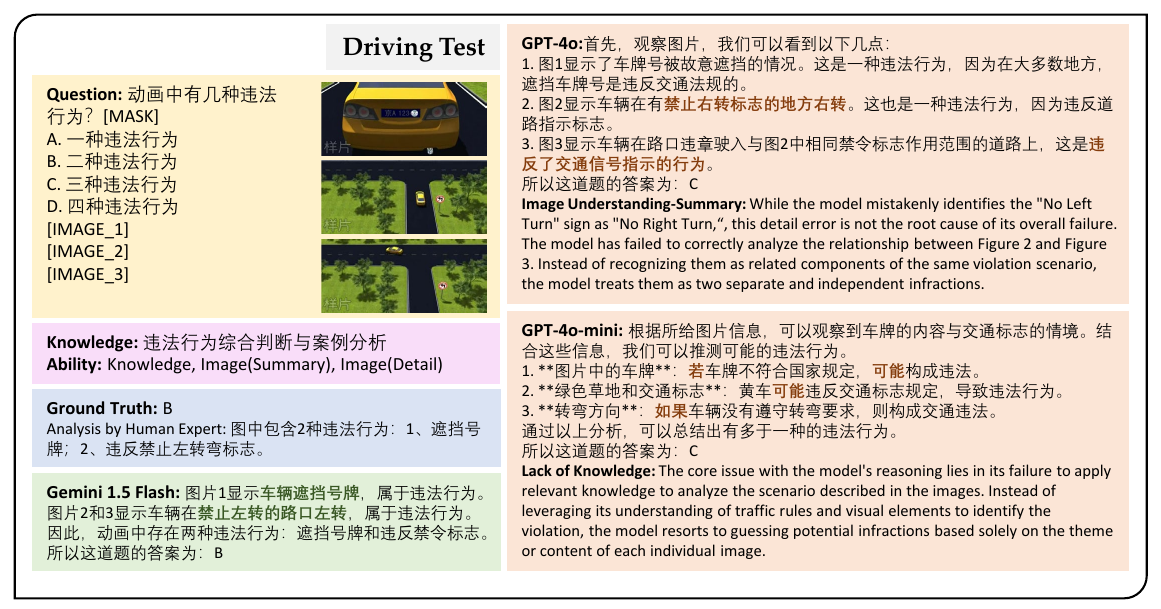}
        \caption{\textbf{Error analysis of an MI-SA question.}}
        \label{figure:case_drive}
    \end{figure}
    
    \begin{figure}[!h]
        \centering
        \includegraphics[width=0.9\textwidth,trim=5 138 5 5,clip]{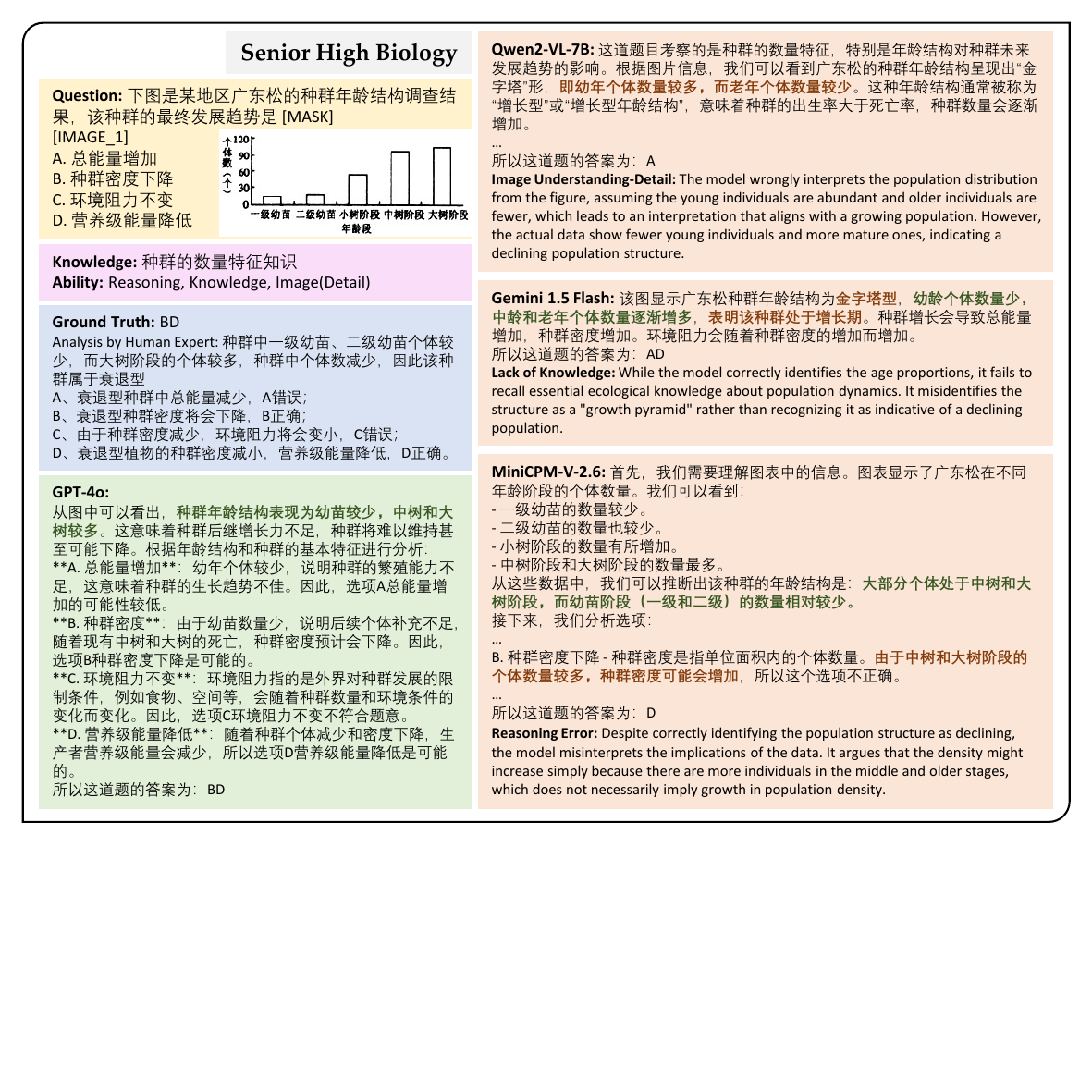}
        \caption{\textbf{Error analysis of an SI-MA question.}}
        \label{figure:case_bio}
    \end{figure}
    
    \begin{figure}[!h]
        \centering
        \includegraphics[width=0.9\textwidth,trim=5 250 5 5,clip]{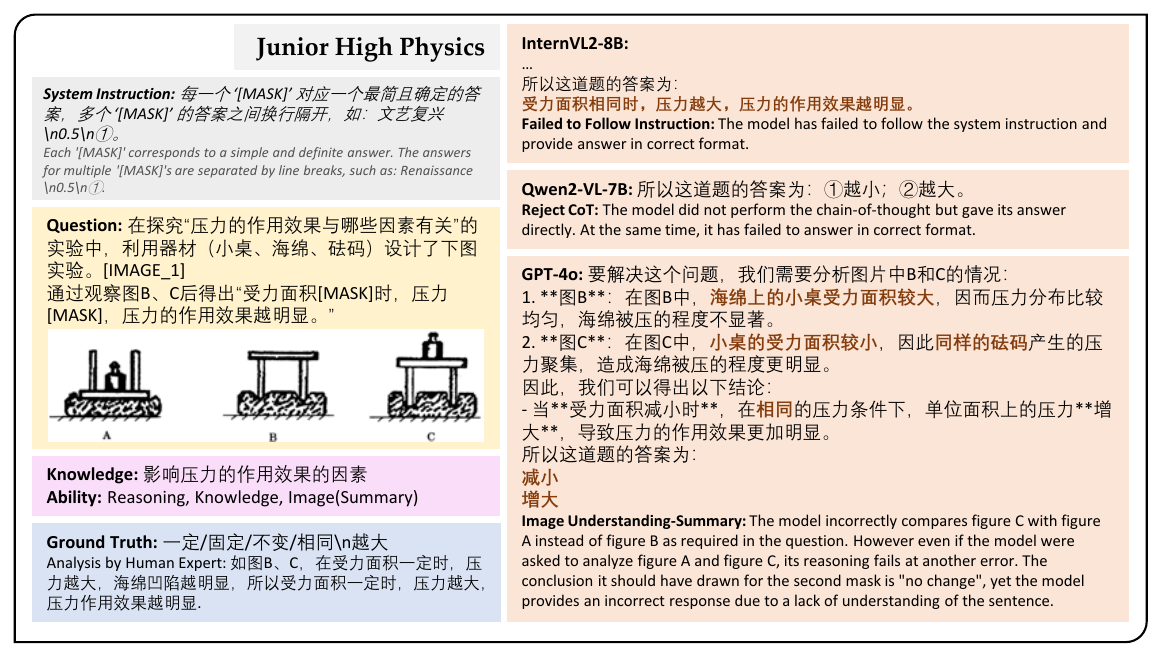}
        \caption{\textbf{Error analysis of an SI-OP question.}}
        \label{figure:case_phy}
    \end{figure}
    
    \begin{figure}[!h]
        \centering
        \includegraphics[width=0.9\textwidth,trim=5 235 5 5,clip]{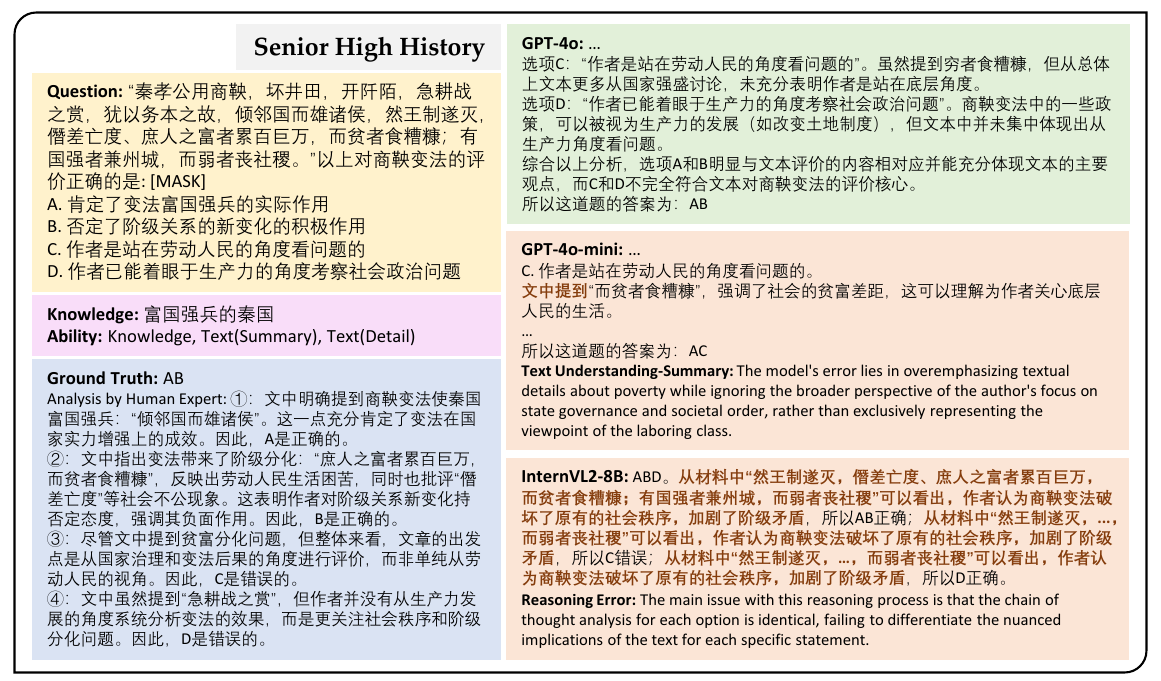}
        \caption{\textbf{Error analysis of an NI-MA question.}}
        \label{figure:case_hist}
    \end{figure}

\end{appendix}

\end{document}